\documentclass[10pt,twocolumn,letterpaper]{article}

\usepackage[pagenumbers]{cvpr} 
\usepackage[accsupp]{axessibility}
\usepackage{graphicx}
\usepackage{amsmath}
\usepackage{amssymb}
\usepackage{booktabs}
\usepackage{multirow}
\usepackage{caption}
\usepackage{enumitem}
\usepackage[table]{xcolor}
\usepackage[labelformat=simple]{subcaption}

\usepackage[pagebackref,breaklinks,colorlinks]{hyperref}

\usepackage[capitalize]{cleveref}
\crefname{section}{Sec.}{Secs.}
\Crefname{section}{Section}{Sections}
\Crefname{table}{Table}{Tables}
\crefname{table}{Tab.}{Tabs.}

\usepackage{pifont}
\newcommand{\cmark}{\ding{51}}%
\newcommand{\xmark}{\ding{55}}%

\def\cvprPaperID{4883} 
\def\confName{CVPR}
\def\confYear{2023}

\begin{document}

\title{Benchmarking Robustness of 3D Object Detection to Common Corruptions \\ in Autonomous Driving}

\author{Yinpeng Dong$^{1,5}$, Caixin Kang$^{2}$, Jinlai Zhang$^{3}$, Zijian Zhu$^{4}$, Yikai Wang$^{1}$, Xiao Yang$^{1}$, \\ Hang Su$^{1,6,7}$, Xingxing Wei$^{2}$, Jun Zhu$^{1,5,6,7}$\thanks{Corresponding author.} \\
  $^{1}$ Dept. of Comp. Sci. and Tech., Institute for AI, Tsinghua-Bosch Joint ML Center,\\
  THBI Lab, BNRist Center, Tsinghua University, Beijing 100084, China \\
  $^{2}$ Institute of Artificial Intelligence, Beihang University, Beijing 100191, China \hspace{0.2ex}
  $^{3}$ Guangxi University \\
  $^{4}$ Institute of Image Communication and Network Engineering, Shanghai Jiao Tong University \\
  $^{5}$ RealAI \hspace{1ex} $^{6}$ Peng Cheng Laboratory \hspace{1ex} $^{7}$ Pazhou Laboratory (Huangpu), Guangzhou, China \\
  \footnotesize{\texttt{\{dongyinpeng,suhangss,dcszj\}@tsinghua.edu.cn, caixinkang@buaa.edu.cn, 1711302013@st.gxu.edu.cn}}
}

\maketitle

\begin{abstract}
3D object detection is an important task in autonomous driving to perceive the surroundings. Despite the excellent performance, the existing 3D detectors lack the robustness to real-world corruptions caused by adverse weathers, sensor noises, etc., provoking concerns about the safety and reliability of autonomous driving systems. To comprehensively and rigorously benchmark the corruption robustness of 3D detectors, in this paper we design 27 types of common corruptions for both LiDAR and camera inputs considering real-world driving scenarios. By synthesizing these corruptions on public datasets, we establish three corruption robustness benchmarks---KITTI-C, nuScenes-C, and Waymo-C. Then, we conduct large-scale experiments on 24 diverse 3D object detection models to evaluate their corruption robustness. Based on the evaluation results, we draw several important findings, including: 1) motion-level corruptions are the most threatening ones that lead to significant performance drop of all models; 2) LiDAR-camera fusion models demonstrate better robustness; 3) camera-only models are extremely vulnerable to image corruptions, showing the indispensability of LiDAR point clouds. We release the benchmarks and codes at \url{https://github.com/kkkcx/3D_Corruptions_AD}.
We hope that our benchmarks and findings can provide insights for future research on developing robust 3D object detection models.
\end{abstract}


\section{Introduction}

As a fundamental task in autonomous driving, 3D object detection aims to identify objects of interest (\eg, vehicles, pedestrians, or cyclists) in the surrounding environment by predicting their categories and the corresponding 3D bounding boxes. 
LiDAR and camera are two important types of sensors for 3D object detection, where the former provides the depth information of road objects as sparse point clouds, while the latter captures abundant semantic information of the scene as color images. Based on the complementary nature of the two modalities, 3D object detection models can be categorized into LiDAR-only~\cite{zhou2018voxelnet,lang2019pointpillars,shi2019pointrcnn,yan2018second,shi2020pv}, camera-only~\cite{mousavian20173d,wang2019pseudo,wang2021fcos3d,wang2022detr3d}, and LiDAR-camera fusion~\cite{chen2017multi,ku2018joint,vora2020pointpainting,liang2018deep} models. 
Since autonomous driving is safety-critical, it is of paramount importance to assess the robustness of 3D object detectors under diverse circumstances before deployed.

Although the recent progress of 3D object detection has led to significant improvements in typical benchmarks (\eg, KITTI~\cite{geiger2012we}, nuScenes~\cite{caesar2020nuscenes}, and Waymo~\cite{sun2020scalability}), 
the existing models based on data-driven deep learning approaches often generalize poorly to the 
corrupted data caused by, \eg, adverse weathers \cite{kilic2021lidar,hahner2021fog,hahner2022lidar}, sensor noises~\cite{carballo2020libre,ren2022benchmarking,hendrycks2018benchmarking}, and uncommon objects~\cite{chan2021segmentmeifyoucan,li2022coda}, posing a formidable obstacle to safe and reliable autonomous driving~\cite{safecar}. To perform robustness evaluation, recent works construct new datasets of road anomalies~\cite{li2022coda,chan2021segmentmeifyoucan,pinggera2016lost,pmlr-v162-hendrycks22a} or under extreme weather conditions~\cite{bijelic2020seeing,pitropov2021canadian,diaz2022ithaca365}. 
Nevertheless, they are usually of small sizes due to the high data collection costs and the rareness of corner cases or adverse weathers.
Other works synthesize common corruptions on clean datasets to benchmark robustness on image classification~\cite{hendrycks2018benchmarking} and point cloud recognition~\cite{ren2022benchmarking,sun2022benchmarking}, but they only consider several simple corruptions, which could be insufficient and unrealistic for 3D object detection. Therefore, it remains challenging to comprehensively characterize different corruptions considering diverse driving scenarios and fairly evaluate corruption robustness of existing models within a unified framework.

\begin{figure*}[t]
  \centering
  \includegraphics[width=0.99\linewidth]{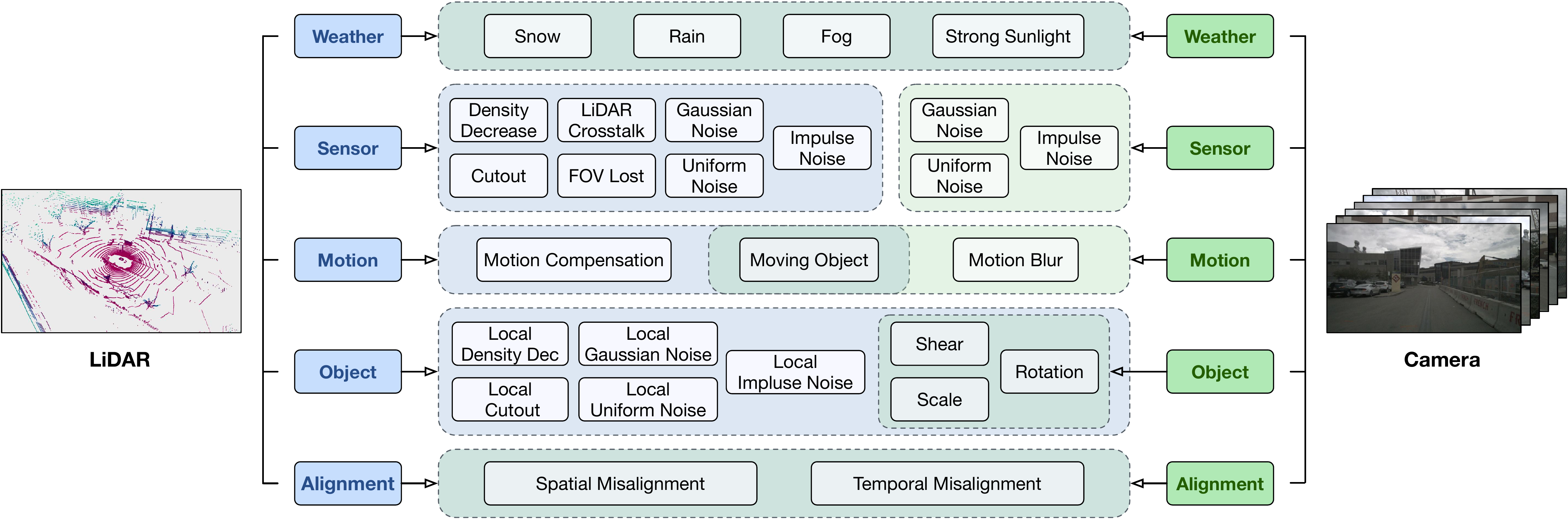}
   \caption{An overview of 27 corruptions for 3D object detection, which are categorized into weather, sensor, motion, object, and alignment levels. As shown, some corruptions are effective for one modality, while the others are applied to both (\eg, \emph{Snow}, \emph{Moving Object}, \emph{Shear}).}
   \vspace{-1ex}
   \label{fig:overview}
\end{figure*}

In this paper, we systematically design \textbf{27} types of common corruptions in 3D object detection for both LiDAR and camera sensors to comprehensively and rigorously evaluate the corruption robustness of current 3D object detectors. The corruptions are grouped into \emph{weather}, \emph{sensor}, \emph{motion}, \emph{object}, and \emph{alignment} levels, covering the majority of real-world corruption cases, as demonstrated in Fig.~\ref{fig:overview}. 
Most of them are specifically designed for autonomous driving (\eg, motion-level ones), which have not been explored before. 
Following~\cite{hendrycks2018benchmarking}, every corruption has five severities, leading to a total number of \textbf{135} distinct corruptions.
By applying them to typical autonomous driving datasets---KITTI~\cite{geiger2012we}, nuScenes~\cite{caesar2020nuscenes}, and Waymo~\cite{sun2020scalability}, we establish three corruption robustness benchmarks---\textbf{KITTI-C}, \textbf{nuScenes-C}, and \textbf{Waymo-C}.
We hope that these large-scale corrupted datasets can serve as general datasets for fairly and comprehensively benchmarking corruption robustness of 3D object  detection models and facilitating future research.

We conduct large-scale experiments to compare the corruption robustness of existing 3D object detection models. Specifically, we evaluate 11 models on KITTI-C, 10 models on nuScenes-C, and 3 models on Waymo-C. The models are of great variety with different input modalities, representation methods, and detection heads. Based on the evaluation results, we find that: 1) the corruption robustness of 3D object detectors is highly correlated with their clean accuracy; 
2) motion-level corruptions impair the model performance most, while being rarely explored before; 3) LiDAR-camera fusion models are more resistant to corruptions, but there is a trade-off between robustness under image corruptions and point cloud corruptions of fusion models. More discussions are provided in Sec.~\ref{sec:conclusion}. Moreover, we study data augmentation strategies~\cite{choi2021part,zhang2022pointcutmix,yun2019cutmix} as potential solutions to improve corruption robustness, but find that they provide a little robustness gain, leaving robustness enhancement of 3D object detection an open problem for future research.

\section{Related Work}

\subsection{3D Object Detection}
\label{sec:3d-detect}
Based on the input modality, we categorize 3D object detection models into LiDAR-only, camera-only, and LiDAR-camera fusion models.

\textbf{LiDAR-only models:} LiDAR point clouds are sparse, irregular, and unordered by nature. To learn useful representations, \emph{voxel-based} methods project point clouds to compact grids. Typically, VoxelNet~\cite{zhou2018voxelnet} rasterizes point clouds into voxels, which are processed by PointNets~\cite{qi2017pointnet} and 3D CNNs. To speed up, SECOND~\cite{yan2018second} introduces sparse 3D convolutions and PointPillars~\cite{lang2019pointpillars} elongates voxels into pillars. Other works exploit information of object parts~\cite{shi2020points} or shape~\cite{zhu2020ssn} to improve the performance. On the other hand, \emph{point-based} methods take raw point clouds as inputs and make predictions on each point. PointRCNN \cite{shi2019pointrcnn} 
proposes a two-stage framework that first generates 3D proposals and then refines the proposals in the canonical coordinates. 3DSSD~\cite{yang20203dssd} is a lightweight one-stage detector with a fusion sampling strategy. To have the best of both worlds, \emph{point-voxel-based} methods are then explored. PV-RCNN \cite{shi2020pv} integrates 3D voxel CNN and PointNet-based set abstraction to efficiently create high-quality proposals.


\textbf{Camera-only models:} 3D object detection based on images is challenging due to the lack of depth information, but attracts extensive attention considering the advantage of low cost. The most straightforward approach is to take \emph{monocular} detection methods~\cite{liu2020smoke,wang2021fcos3d,mousavian20173d,wang2019pseudo,chen2016monocular} and apply post-processing across cameras. For example, Mono3D~\cite{chen2016monocular} generates 3D object proposals scored by semantic features.
SMOKE~\cite{liu2020smoke} combines a single keypoint estimation with regressed 3D variables.
To address the limitation of post-processing in monocular methods, \emph{multi-view} methods fuse information from all cameras in the intermediate layers. DETR3D~\cite{wang2022detr3d} adopts a transformer-based detector~\cite{carion2020end} that fetches the image features by projecting object queries onto images. BEVFormer~\cite{li2022bevformer} exploits spatial-temporal information from multi-view images based on BEV queries.



\textbf{LiDAR-camera fusion models:} To leverage the complementary information from LiDAR and camera inputs, fusion methods are also extensively studied. Following~\cite{liu2022bevfusion}, we classify the newly developed methods into \emph{point-level}, \emph{proposal-level}, and \emph{unified representation} fusion methods. Point-level methods augment LiDAR point clouds with semantic image features and then apply existing LiDAR-only models for 3D detection, including PointPainting \cite{vora2020pointpainting}, EPNet~\cite{huang2020epnet}, PointAugmenting \cite{wang2021pointaugmenting}, Focals Conv~\cite{chen2022focal}, \etc. Proposal-level fusion methods~\cite{chen2017multi,qi2018frustum} generate 3D object proposals and integrate image features into these proposals. FUTR3D~\cite{chen2022futr3d} and TransFusion~\cite{bai2022transfusion} employ a query-based transformer decoder, which fuses image features with object queries. Moreover, BEVFusion~\cite{liu2022bevfusion} unifies the image feature and point cloud feature in a BEV representation space, which stands out as a new fusion strategy.

\subsection{Robustness Benchmarks}\vspace{-0.3ex}

It is well-known that deep learning models lack the robustness to adversarial examples~\cite{szegedy2013intriguing,goodfellow2014explaining}, common corruptions~\cite{hendrycks2018benchmarking}, and other kinds of distribution shifts~\cite{hendrycks2021many,geirhos2018generalisation,geirhos2018imagenet}. In autonomous driving, many works collect new datasets to evaluate model robustness under different conditions. For example, the Seeing Through Fog (STF)~\cite{bijelic2020seeing}, Canadian Adverse Driving Conditions (CADC)~\cite{pitropov2021canadian}, and Ithaca365~\cite{diaz2022ithaca365} datasets are collected in adverse weathers; and others gather road anomalies of 2D images~\cite{li2022coda,chan2021segmentmeifyoucan,pinggera2016lost,pmlr-v162-hendrycks22a}.
Despite the efforts, these datasets only cover limited scenarios due to the high collection costs of rare data. Moreover, as mainly used for evaluation, these datasets have a big domain gap from the large-scale training datasets since they were collected in different cities with varying vehicles and sensors, making it hard for us to examine the effects of different factors (\eg, weather \vs city) on model robustness.

One promising direction is to synthesize real-world corruptions on clean datasets to benchmark model robustness. For example, ImageNet-C~\cite{hendrycks2018benchmarking} is first introduced in image classification with 15 corruption types, ranging from noise, blur, weather to digital corruptions. The similar methodology is further applied to 2D object detection~\cite{michaelis2019benchmarking} and point cloud recognition~\cite{sun2022benchmarking,ren2022benchmarking}. However, many of these studied corruptions are hypothetical and thus unrealistic in the scenario of autonomous driving. It is still challenging to build a comprehensive benchmark for robustness evaluation of 3D object detection considering diverse real-world driving cases. 
We notice that two concurrent works~\cite{li2022common,yu2022benchmarking} to ours also study robustness of 3D object detection in autonomous driving. 
However, they mainly consider specific kinds of 3D detection models (\ie, LiDAR-only models in~\cite{li2022common} and fusion models in~\cite{yu2022benchmarking}) and include limited types of corruptions with less evaluations, as compared in Appendix~A.2.

\section{Corruptions in 3D Object Detection}

\begin{figure*}[t]
  \centering
  \includegraphics[width=0.99\linewidth]{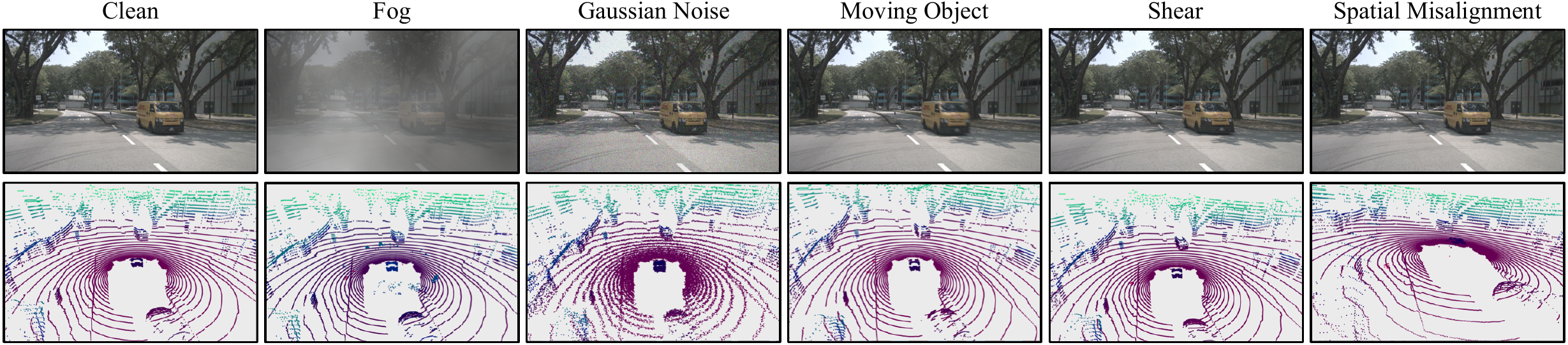}
  \vspace{-1ex}
   \caption{Visualization of typical corruption types of each level in our benchmark (best viewed when zoomed in). Full visualization results of all corruptions are shown in Appendix A.3. }
   \vspace{-1.5ex}
   \label{fig:vis}
\end{figure*}


Real-world corruptions arise from diverse scenarios in autonomous driving, based on which we systematically categorize the corruptions into \emph{weather}, \emph{sensor}, \emph{motion}, \emph{object}, and \emph{alignment} levels. 
We identify common corruption types for each level considering real-world driving scenarios, resulting in \textbf{27} distinct corruptions in total, as shown in Fig.~\ref{fig:overview}.
Among them, some corruptions are applied to both modalities simultaneously, such as weather-level ones, while the others are designed for a single modality, such as sensor-level ones.
We visualize a subset of corruptions in Fig.~\ref{fig:vis}.

\textbf{Weather-level corruptions:} Weather change is usually encountered in autonomous driving, which can dramatically disrupt both LiDAR and camera inputs. For example, \emph{fog} reduces the visibility of objects in images and causes scattered points
due to attenuation and backscattering
\cite{bijelic2020seeing,hahner2021fog,zang2019impact}. Consequently, 3D detectors trained on data collected in normal weather tend to perform poorly under adverse weathers~\cite{bijelic2020seeing}. To study the robustness under weather changes, we consider 4 weather-level corruptions: \emph{Snow}, \emph{Rain}, \emph{Fog}, and \emph{Strong Sunlight}, as they are more common~\cite{bijelic2020seeing,pitropov2021canadian,diaz2022ithaca365}. For LiDAR, we adopt physically based methods~\cite{kilic2021lidar,hahner2022lidar,hahner2021fog} to simulate the effects of rain, snow, and fog on point clouds from normal weather. We simulate the effect of strong sunlight by applying strong Gaussian noises to points along the sun direction~\cite{carballo2020libre}. For camera, we apply image augmentations~\cite{hendrycks2018benchmarking}
to simulate visually realistic weathers.


\textbf{Sensor-level corruptions:} The sensors, when affected by numerous internal or external factors (\eg, sensor vibration~\cite{schlager2022automotive}, lighting conditions~\cite{li2022bevformer,hendrycks2018benchmarking} and reflective materials), can induce various kinds of corruptions to the captured data. 
Based on prior discussions on sensor noises~\cite{carballo2020libre,ren2022benchmarking,berger2014state,hendrycks2018benchmarking}, we design 10 practical sensor-level corruptions---7 for point clouds and 3 for images. 
The point cloud corruptions are: \emph{Density Decrease}, \emph{Cutout}, \emph{LiDAR Crosstalk}, \emph{FOV Lost}, \emph{Gaussian Noise}, \emph{Uniform Noise}, and \emph{Impulse Noise}.
Density decrease simulates missing points commonly observed in typical datasets~\cite{geiger2012we}.
Cutout occurs when laser pulses have no echo in a local region (\eg, puddle) and is simulated by dropping points in a randomly selected area. 
LiDAR crosstalk~\cite{brinon2021methodology} happens when multiple LiDARs operate at close range, which is simulated by applying strong Gaussian noises to a small subset of points. 
FOV lost simulates a limited field-of-view of LiDAR caused by occlusion.
Moreover, due to the ranging inaccuracy of LiDAR, we consider 3 noise corruptions that apply Gaussian, uniform, and impulse noises to point coordinates, respectively. The 3 image corruptions include: \emph{Gaussian Noise}, \emph{Uniform Noise}, and \emph{Impulse Noise} to simulate the visual noise patterns due to low-lighting conditions or defects of camera~\cite{hendrycks2018benchmarking}. Although we design sensor-level corruptions for LiDAR and camera separately, they can occur for both sensors at the same time, affecting LiDAR-camera fusion models further. 

\textbf{Motion-level corruptions:} An autonomous vehicle will encounter several types of corruptions during driving. In this paper, we introduce 3 motion-level corruptions: \emph{Motion Compensation}, \emph{Moving Object}, and \emph{Motion Blur}, which are practical in the real world and studied for the first time. Vehicle ego-motion induces distortions to point clouds since 
the points in a frame are not obtained in the same coordinate system~\cite{zhang2014loam}. To obtain accurate point clouds, motion compensation is typically used with the aid of the localization information~\cite{geiger2012we,caesar2020nuscenes}. However, this process can introduce noises, which we call motion compensation corruption, simulated by adding small Gaussian noises to the rotation and translation matrices of the vehicle's ego pose. The moving object corruption 
denotes the case that an object is moving rapidly in the scene. It can cause shifting points within the object's 3D bounding box~\cite{wen2019gnss} and blur the image patch of the object. The last corruption is motion blur on camera images, which is caused by driving too fast.


\textbf{Object-level corruptions:} Objects in the real world always come in a variety of shapes and materials~\cite{chan2021segmentmeifyoucan,li2022coda}, making it challenging to correctly recognize them. The viewing direction can also lead to wrong recognition of objects~\cite{dong2022viewfool}. Based on this, we introduce 8 object-level corruptions:
\emph{Local Density Decrease}, \emph{Local Cutout}, \emph{Local Gaussian Noise}, \emph{Local Uniform Noise}, \emph{Local Impluse Noise}, \emph{Shear}, \emph{Scale}, and \emph{Rotation}.
The first five corruptions are only applied to LiDAR point clouds to simulate the distortions caused by different object materials or occlusion. As their names indicate, these corruptions only make changes to local sets of points within the objects' 3D bounding boxes.
The last three corruptions simulate shape deformation of objects, and \emph{Rotation} can also simulate different view directions of objects. They can affect both LiDAR and camera inputs. To make consistent distortions to two modalities, we apply the same transformation of shear, scale, or rotation to both points and image patches belonging to the objects in the scene. 


\textbf{Alignment-level corruptions:} It was typically assumed that LiDAR and camera inputs are well aligned before feeding to the fusion models. However, this assumption can be invalid during long-time driving, \eg, the collection of the ONCE dataset~\cite{mao2021one} needs re-calibration almost every day to avoid misalignment between different sensors. In practice, an autonomous vehicle can encounter \emph{Spatial Misalignment} and \emph{Temporal Misalignment}~\cite{yu2022benchmarking}. Spatial misalignment can be caused by sensor vibration due to bumps of the vehicle. We simulate it by adding random noises to the calibration matrices. Temporal misalignment happens when the data is stuck or delayed for a sensor. We keep the input of one modality the same as that at the previous timestamp to simulate temporal misalignment between the two modalities.

\textbf{Discussion about the gap between synthetic and real-world corruptions.} Real-world corruptions can come from multiple and diverse sources. For example, an autonomous vehicle can encounter adverse weather and uncommon objects at the same time, leading to much more complicated corruptions. Although it is impossible to enumerate all real-world corruptions, 
we systematically design 27 corruption types grouped into five levels, which can serve as a practical testbed to perform controllable robustness evaluation.
Especially, for weather-level corruptions, we adopt the state-of-the-art methods for simulation, which are shown to approximate real data well~\cite{hahner2022lidar,hahner2021fog}.
Although there inevitably exists a gap, we validate that the model performance on synthetic weathers are consistent with that on real data under adverse weathers. More discussions are provided in Appendix A.4.

\vspace{-0.5ex}
\section{Corruption Robustness Benchmarks}
\vspace{-0.5ex}

\begin{table*}[t]
\begin{subtable}{0.49\linewidth}
  \centering
 \footnotesize
  \begin{tabular}{c|c|c|c}
  \hline
Model & Modality & Representation & Detection \\
\hline\hline
SECOND~\cite{yan2018second} & LiDAR-only & voxel-based & one-stage \\
PointPillars~\cite{lang2019pointpillars} & LiDAR-only & voxel-based & one-stage \\
PointRCNN~\cite{shi2019pointrcnn} & LiDAR-only & point-based & two-stage \\
Part-$A^2$~\cite{shi2020points} & LiDAR-only & voxel-based & two-stage \\
PV-RCNN~\cite{shi2020pv} & LiDAR-only & point-voxel-based & two-stage \\
3DSSD~\cite{yang20203dssd} & LiDAR-only & point-based & one-stage \\
\hline
SMOKE~\cite{liu2020smoke} & camera-only & monocular & one-stage \\
PGD~\cite{wang2022probabilistic} & camera-only & monocular & one-stage \\
ImVoxelNet~\cite{rukhovich2022imvoxelnet} & camera-only & monocular & one-stage \\
\hline
EPNet~\cite{huang2020epnet} & fusion & point-level & two-stage \\
Focals Conv~\cite{chen2022focal} & fusion & point-level & two-stage \\
\hline
  \end{tabular}
  \caption{Evaluated models on KITTI-C.}\label{tab:kitti-c}
  \end{subtable}
  \hspace{1ex}
  \begin{subtable}{0.48\linewidth}
\centering
 \footnotesize
  \begin{tabular}{c|c|c|c}
  \hline
Model & Modality & Representation & Detection \\
\hline\hline
PointPillars~\cite{lang2019pointpillars} & LiDAR-only & voxel-based & one-stage \\
SSN~\cite{zhu2020ssn} & LiDAR-only & voxel-based & one-stage \\
CenterPoint~\cite{yin2021center} & LiDAR-only & voxel-based & two-stage \\
\hline
FCOS3D~\cite{wang2021fcos3d} & camera-only & monocular & one-stage \\
PGD~\cite{wang2022probabilistic} & camera-only & monocular & one-stage \\
DETR3D~\cite{wang2022detr3d} & camera-only & multi-view & query-based \\
BEVFormer~\cite{li2022bevformer} & camera-only & multi-view & query-based \\
\hline
FUTR3D~\cite{chen2022futr3d} & fusion & proposal-level & query-based \\
TransFusion~\cite{bai2022transfusion} & fusion & proposal-level & query-based \\
BEVFusion~\cite{liu2022bevfusion} & fusion & unified & query-based \\
\hline
   \end{tabular}
   \vspace{1.2ex}
   \caption{Evaluated models on nuScenes-C.}\label{tab:nuscene-c}
  \end{subtable}
  \caption{The 3D object detection models adopted for corruption robustness evaluation on KITTI-C and nuScenes-C. We show the input modality, representation learning method (see Sec.~\ref{sec:3d-detect}), and detection head of each model.}
   \label{tab:model-summary}
\end{table*}

To comprehensively evaluate the corruption robustness of 3D object detection models, we establish three corruption robustness benchmarks based on the most widely used datasets in autonomous driving---KITTI~\cite{geiger2012we}, nuScenes~\cite{caesar2020nuscenes}, and Waymo~\cite{sun2020scalability}. We apply the aforementioned corruptions to the validation sets of these datasets and obtain \textbf{KITTI-C}, \textbf{nuScenes-C}, and \textbf{Waymo-C}, respectively. Note that although several corruptions naturally appear in few samples of the datasets, we 
still apply the synthetic corruptions to all data to fairly compare model robustness under different corruptions and reduce the efforts of filtering data. Besides, we build a unified toolkit comprising of all corruptions, that can be used for other datasets as well. Below we introduce the dataset details, evaluation metrics, and evaluated models of the three benchmarks, respectively.

\subsection{KITTI-C} 
The KITTI dataset~\cite{geiger2012we} contains 3712 training, 3769 validation, and 7518 test samples. As we do not have access to the test set, KITTI-C is constructed upon the validation set. 
Among the corruptions, we do not include \emph{FOV Lost}, \emph{Motion Compensation} and \emph{Temporal Misalignment} since: 1) 3D object detection models usually take front-view point clouds of $90^{\circ}$ FOV as inputs since the KITTI dataset only provides box annotations in front of the vehicle; 2) the localization and timestamp information of each frame is not provided in the dataset. Therefore, there are 24 corruptions in KITTI-C with 5 severities for each following~\cite{hendrycks2018benchmarking}.

The standard evaluation is performed on \emph{Car}, \emph{Pedestrian} and \emph{Cyclist} categories at \emph{Easy}, \emph{Moderate} and \emph{Hard} levels of difficulty. 
The evaluation metric is the Average Precision (AP) with 40 recall positions at an IoU threshold 0.7 for cars and 0.5 for pedestrians/cyclists. We denote model performance on the original validation set as $\mathrm{AP_{clean}}$. For each corruption type $c$ at each severity $s$, we adopt the same metric to measure model performance as $\mathrm{AP}_{c,s}$. Then, the \emph{corruption robustness} of a model is calculated by averaging over all corruption types and severities as
\begin{equation}\small
    \mathrm{AP_{cor}} = \frac{1}{|\mathcal{C}|}\sum_{c\in \mathcal{C}} \frac{1}{5}\sum_{s=1}^{5}\mathrm{AP}_{c,s},
\end{equation}
where $\mathcal{C}$ is the set of corruptions in evaluation. Note that for different kinds of 3D object detectors, the set of corruptions can be different (\eg, we do not evaluate camera noises for LiDAR-only models), thus the results of $\mathrm{AP_{cor}}$ are \emph{not} directly comparable between different kinds of models and we perform a fine-grained analysis under each corruption.
We also calculate \emph{relative corruption error (RCE)} by measuring the percentage of performance drop as
\begin{equation}
\small\label{eq:2}
\mathrm{RCE}_{c,s}=\frac{\mathrm{AP_{clean}}-\mathrm{AP}_{c,s}}{\mathrm{AP_{clean}}};\;\mathrm{RCE}=\frac{\mathrm{AP_{clean}}-\mathrm{AP_{cor}}}{\mathrm{AP_{clean}}}.
\end{equation}

We select 11 representative 3D object detection models trained on KITTI, including 6 LiDAR-only models: \textit{SECOND}~\cite{yan2018second}, \textit{PointPillars}~\cite{lang2019pointpillars}, \textit{PointRCNN}~\cite{shi2019pointrcnn}, \textit{Part-$A^2$} \cite{shi2020points},
\textit{PV-RCNN}~\cite{shi2020pv}, and \textit{3DSSD}~\cite{yang20203dssd}; 3 camera-only models: \textit{SMOKE}~\cite{liu2020smoke}, \textit{PGD}~\cite{wang2022probabilistic}, and \textit{ImVoxelNet}~\cite{rukhovich2022imvoxelnet}; and 2 LiDAR-camera fusion models: \textit{EPNet}~\cite{huang2020epnet} and \textit{Focals Conv} \cite{chen2022focal}. The details regarding their representations and detection heads are shown in Table~\ref{tab:kitti-c}. 

\subsection{nuScenes-C}

The nuScenes dataset~\cite{caesar2020nuscenes} contains 1000 sequences of approximately 20s duration with a LiDAR frequency of 20 FPS. The box annotations are provided for every 0.5s. Each frame has one point cloud and six images covering $360^{\circ}$ horizontal FOV. In total, there are 40k frames which are split into 28k, 6k, 6k for training, validation, and testing. 
As the dataset provides full annotations and information of vehicle pose and timestamp, we can simulate all corruptions. Thus, we apply all 27 corruptions to the nuScenes validation set with 5 severities to obtain nuScenes-C. 

For 3D object detection, the main evaluation metrics are mean Average Precision (mAP) and nuScenes detection score (NDS) computed on 10 object categories. The mAP is calculated using the 2D center distance on the ground plane instead of the 3D IoU. The NDS metric consolidates mAP and other aspects (\eg, scale, orientation) into a unified score. Similar to KITTI-C, we denote model performance on the validation set as $\mathrm{mAP_{clean}}$ and $\mathrm{NDS_{clean}}$, and measure the corruption robustness $\mathrm{mAP_{cor}}$ and $\mathrm{NDS_{cor}}$ by averaging over all corruptions and severities. We also compute the relative corruption error $\mathrm{RCE}$ under both mAP and NDS metrics similar to Eq.~\eqref{eq:2}.

On nuScenes-C, we select 10 3D detectors, including 3 LiDAR-only models: \emph{PointPillars}~\cite{lang2019pointpillars}, \emph{SSN}~\cite{zhu2020ssn}, and \emph{CenterPoint}~\cite{yin2021center}; 4 camera-only models: \emph{FCOS3D}~\cite{wang2021fcos3d}, \emph{PGD} \cite{wang2022probabilistic}, \emph{DETR3D}~\cite{wang2022detr3d}, and \emph{BEVFormer}~\cite{li2022bevformer}; and 3 LiDAR-camera fusion models: \emph{FUTR3D}~\cite{chen2022futr3d}, \emph{TransFusion}~\cite{bai2022transfusion}, and \emph{BEVFusion}~\cite{liu2022bevfusion}. The model details are shown in Table~\ref{tab:nuscene-c}.

\begin{table*}[!t]
\footnotesize
\setlength{\tabcolsep}{2.8pt}
\begin{center}
\begin{tabular}{c|c|cccccc|ccc|cc}
\hline
\multicolumn{2}{c|}{\multirow{2}{*}{\textbf{Corruption}}} & \multicolumn{6}{c|}{\textbf{LiDAR-only}} &  \multicolumn{3}{c|}{\textbf{Camera-only}} & \multicolumn{2}{c}{\textbf{LC Fusion}} \\
\multicolumn{2}{c|}{} & SECOND & PointPillars & PointRCNN & Part-$A^2$ & PV-RCNN & 3DSSD & SMOKE & PGD & ImVoxelNet & EPNet & Focals Conv \\
\hline\hline
\rowcolor{lightgray}\multicolumn{2}{c|}{\textbf{None} ($\mathrm{AP_{clean}}$)} & 81.59 & 78.41 & 80.57 & 82.45 & \bf84.39 & 80.03 & 7.09 & 8.10 & \bf11.49 & 82.72 & \bf85.88 \\
\hline
\multirow{4}{*}{\textbf{Weather}} & Snow & \bf52.34 & 36.47 & 50.36 & 42.70 & \bf52.35 & 27.12 & 2.47 & 0.63 & 0.22 & 34.58 & 34.77 \\
& Rain & \bf52.55 & 36.18 & 51.27 & 41.63 & 51.58 & 26.28 & 3.94 & 3.06 & 1.24 & 36.27 & 41.30 \\
& Fog & 74.10 & 64.28 & 72.14 & 71.61 & \bf79.47 & 45.89 & 5.63 & 0.87 & 1.34 & 44.35 & 44.55 \\
& Sunlight & 78.32 & 62.28 & 62.78 & 76.45 & 79.91 & 26.09 & 6.00 & 7.07 & 10.08 & 69.65 & \bf80.97 \\
\hline
\multirow{9}{*}{\textbf{Sensor}} & Density & 80.18 & 76.49 & 80.35 & 80.53 & 82.79 & 77.65 & - & - & - & 82.09 & \bf84.95 \\
& Cutout & 73.59 & 70.28 & 73.94 & 76.08 & 76.09 & 73.05 & - & - & - & 76.10 & \bf78.06 \\
& Crosstalk & 80.24 & 70.85 & 71.53 & 79.95 & 82.34 & 46.49 & - & - & - & 82.10 & \bf85.82 \\
& Gaussian (L) & 64.90 & 74.68 & 61.20 & 60.73 & 65.11 & 59.14 & - & - & - & 60.88 & \bf82.14 \\
& Uniform (L) & 79.18 & 77.31 & 76.39 & 77.77 & 81.16 & 74.91 & - & - & - & 79.24 & \bf85.81 \\
& Impulse (L) & 81.43 & 78.17 & 79.78 & 80.80 & 82.81 & 78.28 & - & - & - & 81.63 & \bf85.01 \\
& Gaussian (C) & - & - & - & - & - & - & 1.56 & 1.71 & 2.43 & 80.64 & \bf80.97 \\
& Uniform (C) & - & - & - & - & - & - & 2.67 & 3.29 & 4.85 & 81.61 & \bf83.38 \\
& Impulse (C) & - & - & - & - & - & - & 1.83 & 1.14 & 2.13 & \bf81.18 & 80.83 \\
\hline
\multirow{2}{*}{\textbf{Motion}} & Moving Obj. & 52.69 & 50.15 & 50.54 & 54.62 & 54.60 & 52.47 & 1.67 & 2.64 & 5.93 & \bf55.78 & 49.14 \\
& Motion Blur & - & - & - & - & - & - & 3.51 & 3.36 & 4.19 & 74.71 & \bf81.08 \\
\hline
\multirow{8}{*}{\textbf{Object}} & Local Density & 75.10 & 69.56 & 74.24 & 79.57 & 77.63 & 77.96 & - & - & - & 76.73 & \bf80.84 \\
& Local Cutout & 68.29 & 61.80 & 67.94 & 75.06 & 72.29 & 73.22 & - & - & - & 69.92 & \bf76.64 \\
& Local Gaussian & 72.31 & 76.58 & 69.82 & 77.44 & 70.44 & 75.11 & - & - & - & 75.76 & \bf82.02 \\
& Local Uniform & 80.17 & 78.04 & 77.67 & 80.77 & 82.09 & 78.64 & - & - & - & 81.71 & \bf84.69 \\
& Local Impulse & 81.56 & 78.43 & 80.26 & 82.25 & 84.03 & 79.53 & - & - & - & 82.21 & \bf85.78 \\
& Shear & 41.64 & 39.63 & 39.80 & 37.08 & \bf47.72 & 26.56 & 1.68 & 2.99 & 1.33 & 41.43 & 45.77 \\
& Scale & 73.11 & 70.29 & 71.50 & 75.90 & \bf76.81 & 75.02 & 0.13 & 0.15 & 0.33 & 69.05 & 69.48 \\
& Rotation & 76.84 & 72.70 & 75.57 & 77.50 & \bf79.93 & 76.98 & 1.11 & 2.14 & 2.57 & 74.62 & 77.76 \\
\hline
\multirow{1}{*}{\textbf{Alignment}} & Spatial & - & - & - & - & - & - & - & - & - & 35.14 & 43.01 \\
\hline
\rowcolor{lightgray}\multicolumn{2}{c|}{\textbf{Average} ($\mathrm{AP_{cor}}$)} & 70.45 & 65.48 & 67.74 & 69.92 & \bf72.59 & 60.55 & 2.68 & 2.42 & \bf3.05 & 67.81 & \bf71.87 \\
\hline
\end{tabular}
\end{center}
\vspace{-3ex}
\caption{The benchmarking results of 11 3D object detectors on \textbf{KITTI-C}. We show the performance under each corruption and  the overall corruption robustness $\mathrm{AP_{cor}}$ averaged over all corruption types. The results are evaluated based on the car class at moderate difficulty.}
\label{tab:results-kitti}
\end{table*}

\subsection{Waymo-C}

The Waymo open dataset \cite{sun2020scalability} consists of 798 scenes for training and 202 scenes for validation. Similar to nuScenes-C, Waymo-C is constructed by applying all 27 corruptions to the Waymo validation set with 5 severities. The official evaluation metrics are mAP and mAPH by taking the heading accuracy into consideration. We similarly calculate the corruption robustness and relative corruption error on Waymo-C. Due to the license agreement, there are no pre-train models publicly. Thus, we train the LiDAR-only \emph{PointPillars}~\cite{lang2019pointpillars}, camera-only \emph{BEVFormer}~\cite{li2022bevformer}, and LiDAR-camera fusion \emph{TransFusion}~\cite{bai2022transfusion} on a subset of training data~\cite{li2022bevformer} for robustness evaluation.

\section{Benchmarking Results}

We present the evaluation results on KITTI-C in Sec.~\ref{sec:5-1}, nuScenes-C in Sec.~\ref{sec:5-2}, and leave the results on Waymo-C in Appendix D. 
We summarize the key findings in Sec.~\ref{sec:conclusion}.

\subsection{Results on KITTI-C}\label{sec:5-1}

\begin{table*}[!t]
\footnotesize
\setlength{\tabcolsep}{3.5pt}
\begin{center}
\begin{tabular}{c|c|ccc|cccc|ccc}
\hline
\multicolumn{2}{c|}{\multirow{2}{*}{\textbf{Corruption}}} & \multicolumn{3}{c|}{\textbf{LiDAR-only}} &  \multicolumn{4}{c|}{\textbf{Camera-only}} & \multicolumn{3}{c}{\textbf{LC Fusion}} \\
\multicolumn{2}{c|}{} & PointPillars & SSN & CenterPoint & FCOS3D & PGD & DETR3D & BEVFormer & FUTR3D & TransFusion & BEVFusion \\
\hline\hline
\rowcolor{lightgray}\multicolumn{2}{c|}{\textbf{None} ($\mathrm{mAP_{clean}}$)} & 27.69 & 46.65 & \bf59.28 & 23.86 & 23.19 & 34.71 & \bf41.65 & 64.17 & 66.38 & \bf68.45 \\
\hline
\multirow{4}{*}{\textbf{Weather}} & Snow & 27.57 & 46.38 & 55.90 & 2.01 & 2.30 & 5.08 & 5.73 & 52.73 & \bf63.30 & 62.84 \\
& Rain & 27.71 & 46.50 & 56.08 & 13.00 & 13.51 & 20.39 & 24.97 & 58.40 & 65.35 & \bf66.13 \\
& Fog & 24.49 & 41.64 & 43.78 & 13.53 & 12.83 & 27.89 & 32.76 & 53.19 & 53.67 & \bf54.10 \\
& Sunlight & 23.71 & 40.28 & 54.20 & 17.20 & 22.77 & 34.66 & 41.68 & 57.70 & 55.14 & \bf64.42 \\
\hline
\multirow{10}{*}{\textbf{Sensor}} & Density & 27.27 & 46.14 & 58.60 & - & - & - & - & 63.72 & 65.77 & \bf67.79 \\
& Cutout & 24.14 & 40.95 & 56.28 & - & - & - & - & 62.25 & 63.66 & \bf66.18 \\
& Crosstalk & 25.92 & 44.08 & 56.64 & - & - & - & - & 62.66 & 64.67 & \bf67.32 \\
& FOV Lost & 8.87 & 15.40 & 20.84 & - & - & - & - & 26.32 & 24.63 & \bf27.17 \\
& Gaussian (L) & 19.41 & 39.16 & 45.79 & - & - & - & - & 58.94 & 55.10 & \bf60.64 \\
& Uniform (L) & 25.60 & 45.00 & 56.12 & - & - & - & - & 63.21 & 64.72 & \bf66.81 \\
& Impulse (L) & 26.44 & 45.58 & 57.67 & - & - & - & - & 63.43 & 65.51 & \bf67.54 \\
& Gaussian (C) & - & - & - & 3.96 & 4.33 & 14.86 & 15.04 & 54.96 & \bf64.52 & 64.44 \\
& Uniform (C) & - & - & - & 8.12 & 8.48 & 21.49 & 23.00 & 57.61 & 65.26 & \bf65.81 \\
& Impulse (C) & - & - & - & 3.55 & 3.78 & 14.32 & 13.99 & 55.16 & \bf64.37 & 64.30 \\
\hline
\multirow{3}{*}{\textbf{Motion}} & Compensation & 3.85 & 10.39 & 11.02 & - & - & - & - & \bf31.87 & 9.01 & 27.57 \\
& Moving Obj. & 19.38 & 35.11 & 44.30 & 10.36 & 10.47 & 16.63 & 20.22 & 45.43 & 51.01 & \bf51.63 \\
& Motion Blur & - & - & - & 10.19 & 9.64 & 11.06 & 19.79 & 55.99 & 64.39 & \bf64.74 \\
\hline
\multirow{8}{*}{\textbf{Object}} & Local Density & 26.70 & 45.42 & 57.55 & - & - & - & - & 63.60 & 65.65 & \bf67.42 \\
& Local Cutout & 17.97 & 32.16 & 48.36 & - & - & - & - & 61.85 & 63.33 & \bf63.41 \\
& Local Gaussian & 25.93 & 43.71 & 51.13 & - & - & - & - & 62.94 & 63.76 & \bf64.34 \\
& Local Uniform & 27.69 & 46.87 & 57.87 & - & - & - & - & 64.09 & 66.20 & \bf67.58 \\
& Local Impulse & 27.67 & 46.88 & 58.49 & - & - & - & - & 64.02 & 66.29 & \bf67.91 \\
& Shear & 26.34 & 43.28 & 49.57 & 17.20 & 16.66 & 17.46 & 24.71 & 55.42 & \bf62.32 & 60.72 \\
& Scale & 27.29 & 45.98 & 51.13 & 6.75 & 6.57 & 12.02 & 17.64 & 56.79 & 64.13 & \bf64.57 \\
& Rotation & 27.80 & 46.93 & 54.68 & 17.21 & 16.84 & 27.28 & 33.97 & 59.64 & 63.36 & \bf65.13 \\
\hline
\multirow{2}{*}{\textbf{Alignment}} & Spatial & - & - & - & - & - & - & - & 63.77 & 66.22 & \bf68.39 \\
& Temporal & - & - & - & - & - & - & - & \bf51.43 & 43.65 & 49.02 \\
\hline
\rowcolor{lightgray}\multicolumn{2}{c|}{\textbf{Average} ($\mathrm{mAP_{cor}}$)} & 23.42 & 40.37 & \bf49.81 & 10.26 & 10.68 & 18.60 & \bf22.79 & 56.99 & 58.73 & \bf61.03 \\
\hline
\end{tabular}
\end{center}
\vspace{-3ex}
\caption{The benchmarking results of 10 3D object detectors on \textbf{nuScenes-C}. We show the performance under each corruption and  the overall corruption robustness $\mathrm{mAP_{cor}}$ averaged over all corruption types.}
\label{tab:results-nuscenes}
\end{table*}

\begin{figure}[t]
  \centering
  \includegraphics[width=0.99\linewidth]{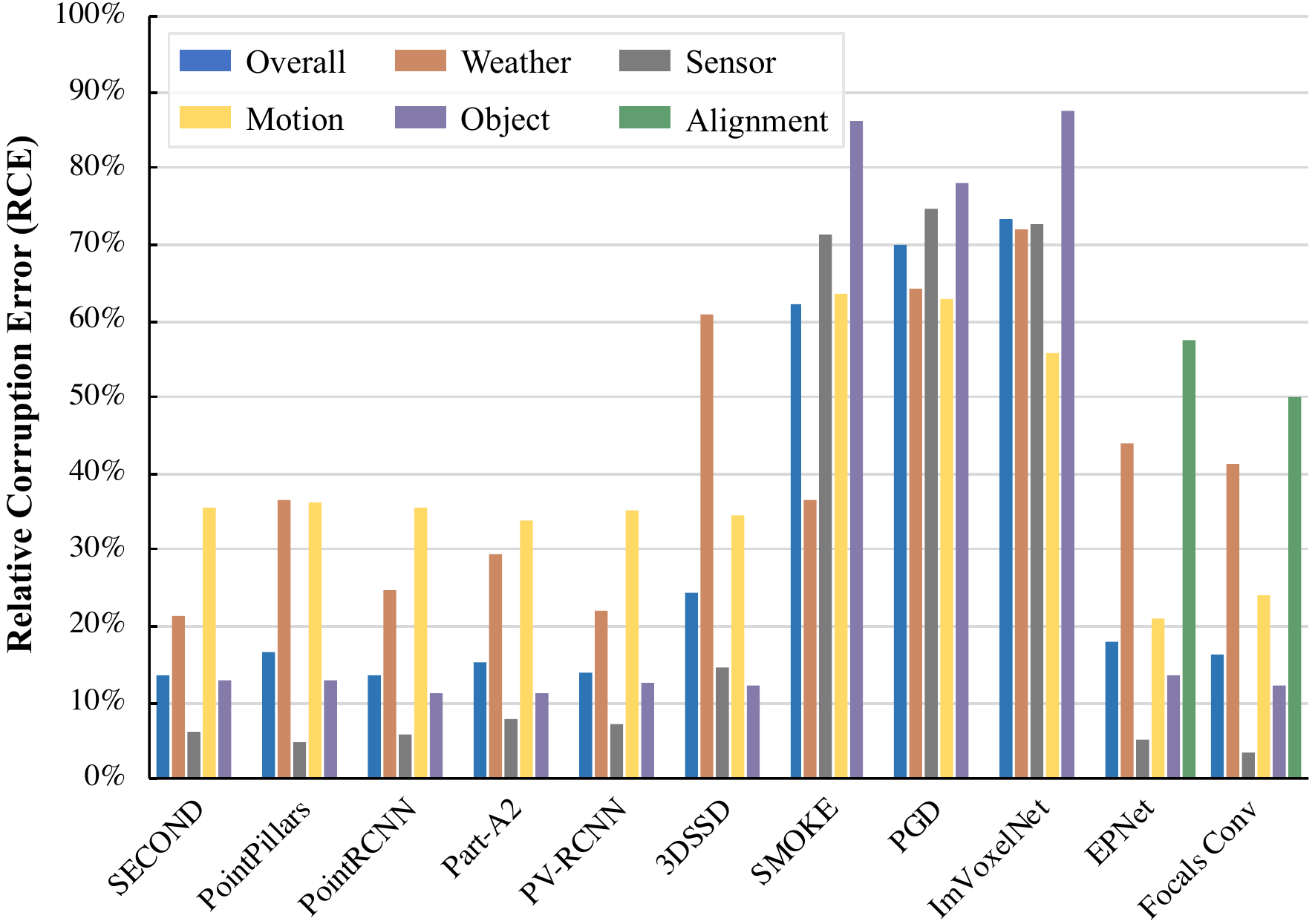}
   \caption{The relative corruption error $\mathrm{RCE}$ of 11 3D object detectors on \textbf{KITTI-C}. We show the overall results under all corruptions and the results under each level of corruptions.}
   \label{fig:plot-kitti}
\end{figure}

We show the corruption robustness of 11 3D object detection models on KITTI-C in Table \ref{tab:results-kitti}, in which we only report the results on the car class at moderate difficulty, while leaving full results of other classes and difficulties in Appendix B. Overall, the corruption robustness is highly correlated with the clean accuracy, as the models (\eg, PV-RCNN, Focals Conv) with higher $\mathrm{AP_{clean}}$ also achieve higher $\mathrm{AP_{cor}}$. It is not surprising due to the consistent performance degradation of different models. We further show the relative corruption error $\mathrm{RCE}$ of these models under each level of corruptions in Fig.~\ref{fig:plot-kitti}. 
Based on the evaluation results, we provide the analyses below.

\textbf{Comparison of corruption types.} Based on Table~\ref{tab:results-kitti} and Fig.~\ref{fig:plot-kitti}, we can observe that weather-level and motion-level corruptions affect the performance of LiDAR-only and fusion models most, while all corruptions cause significant performance drop for camera-only models.  For example, \emph{Snow} and \emph{Rain} lead to more than $35\%$ $\mathrm{RCE}$ for all models, demonstrating the threats of adverse weathers on 3D object detectors. Besides, \emph{Moving Object} and \emph{Shear} are also challenging for all models, while  \emph{Spatial Misalignment} has a great impact on fusion models. On the other hand, most models exhibit negligible performance drop under sensor-level and object-level corruptions, mainly due to their ubiquity in the training dataset.

\begin{figure}[t]
  \centering
  \includegraphics[width=0.99\linewidth]{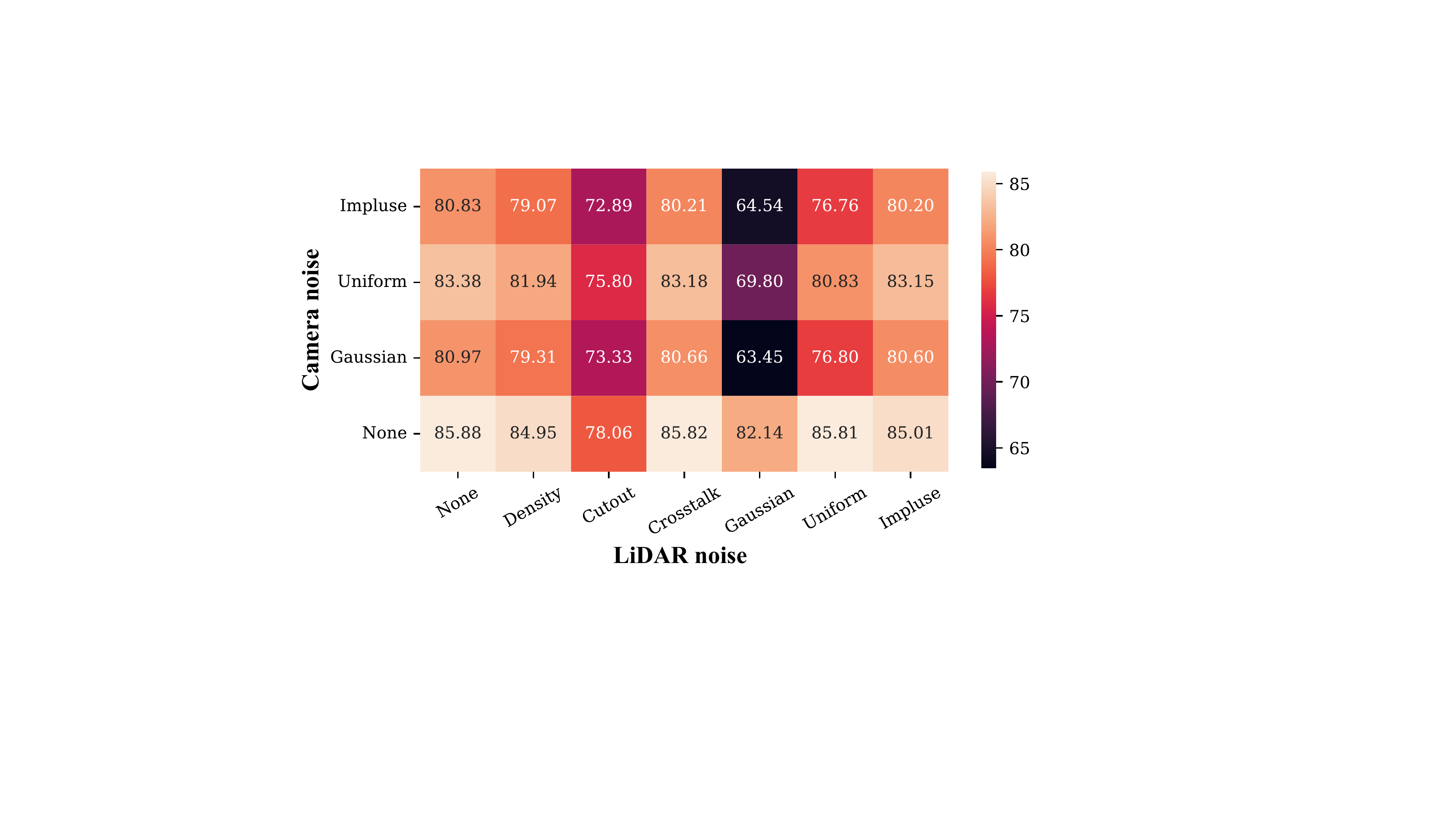}
   \caption{The performance of Focals Conv~\cite{chen2022focal} under the concurrence of LiDAR and camera noises.}
   \label{fig:heatmap}
\end{figure}

\textbf{Comparison of 3D object detectors.} Due to the inferior performance of camera-only models, we mainly compare LiDAR-only and LiDAR-camera fusion models. We notice that for corruptions that affect both modalities (\eg, \emph{Snow}, \emph{Moving Object}, \emph{Shear}), LiDAR-only models lead to better performance. But for those that only corrupt point clouds (\eg, sensor noises), fusion models are more competitive. This is due to that the accurate image data can endow fusion models with better robustness under point cloud noises, but when images are also corrupted, fusion models are affected by both inputs, resulting in inferior performance. To further validate this, we apply sensor noises to LiDAR and camera inputs at the same time. We show the performance of Focals Conv~\cite{chen2022focal} under the concurrence of LiDAR and camera noises in Fig.~\ref{fig:heatmap}. It can be seen that the accuracy of Focals Conv further drops in the presence of both LiDAR and camera noises, leading to worse performance than LiDAR-only models that cannot be affected by camera noises. The results demonstrate that although fusion models are more robust to noises of one modality, they are potentially exposed to corruptions from multiple sensors.

\textbf{Comparison of LiDAR-only models.} Among the six LiDAR-only detectors, we find that SECOND~\cite{yan2018second}, PointRCNN~\cite{shi2019pointrcnn}, and PV-RCNN~\cite{shi2020pv} possess better relative corruption robustness than the others, whose $\mathrm{RCE}$ is $13.65\%$, $13.61\%$, and $13.99\%$. The worst model is 3DSSD, exhibiting a $24.34\%$ performance drop. In general, there does not exist a clear margin of robustness between voxel-based and point-based detectors, or between one-stage and two-stage detectors, different from previous findings~\cite{li2022common}.  However, we notice that the worst two models PointPillars~\cite{lang2019pointpillars} and 3DSSD~\cite{yang20203dssd} are developed for improving the efficiency of 3D object detection, which may indicate a trade-off between corruption robustness and efficiency.



\subsection{Results on nuScenes-C}\label{sec:5-2}

\begin{figure}[t]
  \centering
  \includegraphics[width=0.99\linewidth]{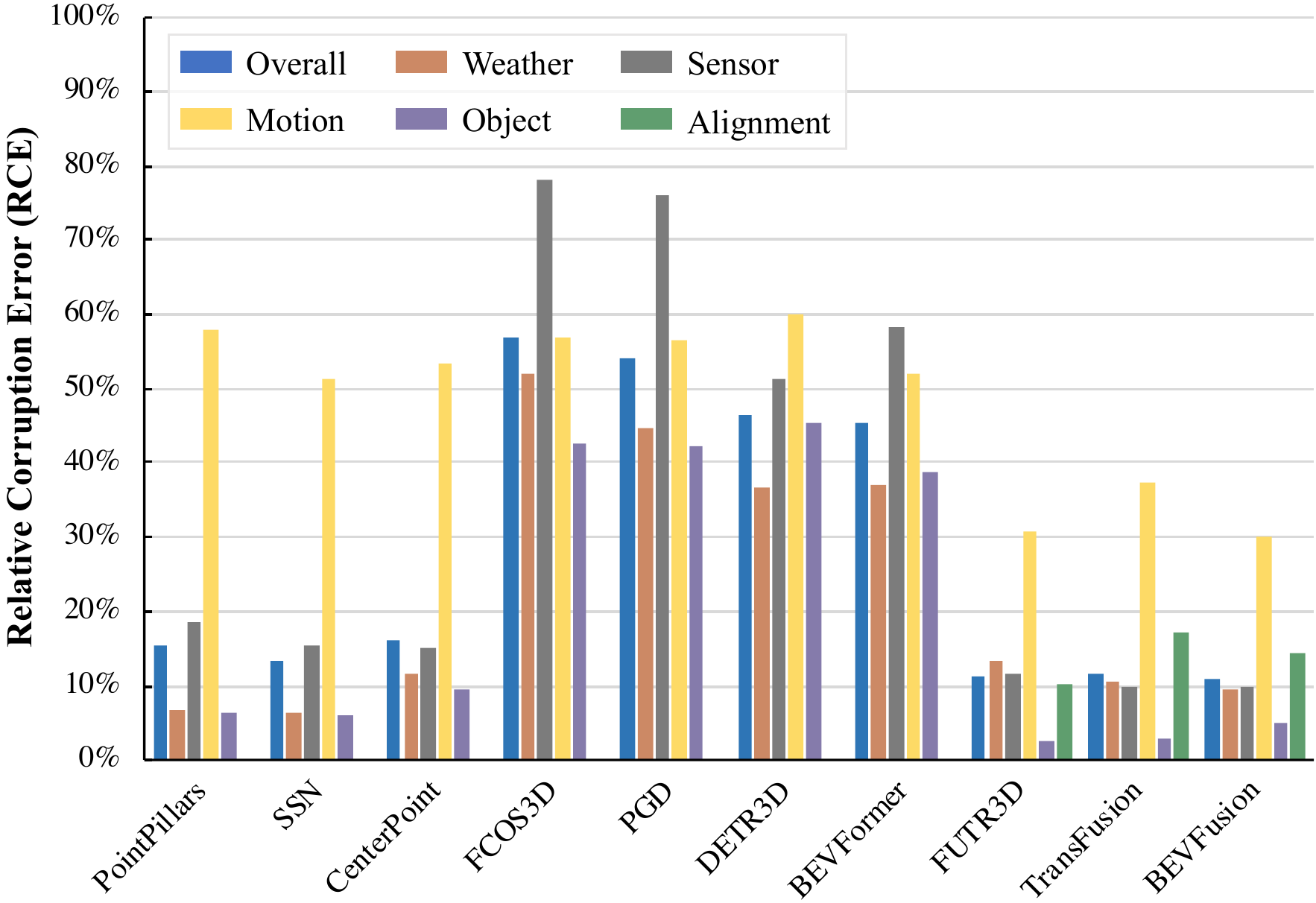}
   \caption{The relative corruption error $\mathrm{RCE}$ of 10 3D object detectors on \textbf{nuScenes-C}. We show the overall results under all corruptions and the results under each level of corruptions.}
   \label{fig:plot-nuscene}
\end{figure}

We report the corruption robustness of 10 3D detectors on nuScenes-C in Table \ref{tab:results-nuscenes} under the mAP metric, and leave the results under the NDS metric in Appendix C. The model performance is consistent for both metrics. We further show the relative corruption error $\mathrm{RCE}$ under each level of corruptions in Fig.~\ref{fig:plot-nuscene}. Similar to the results on KITTI-C, models that have higher clean accuracy generally achieve better corruption robustness. 
But differently, the nuScenes dataset provides multi-view images, thus the camera-only models achieve competitive clean accuracy with  LiDAR-only models, enabling us to compare their performance.
We provide more detailed analyses below. 

\textbf{Comparison of corruption types.} From Fig.~\ref{fig:plot-nuscene}, we can observe that motion-level corruptions are significantly more detrimental to LiDAR-only and LiDAR-camera fusion models. 
They give rise to more than $50\%$ performance drop for LiDAR-only models and about $30\%$ drop for fusion models. 
Similar to KITTI-C, all corruptions remarkably degrade the performance of camera-only models. A notable difference from KITTI-C is that most models are resistant to weather-level corruptions. We think that the adverse weathers (\eg, rain) contained in the nuScenes dataset enable the detectors to predict robustly under weather-level corruptions. Among all corruptions, \emph{FOV Lost} 
and \emph{Motion Compensation} impair the models most, mainly due to the large distortions of the LiDAR point clouds. 

\textbf{Comparison of 3D object detectors.}
For different categories of 3D object detectors, camera-only models are more prone to common corruptions, whose performance drops more than $40\%$ under $\mathrm{RCE}$. On the contrary, LiDAR-only and fusion models exhibit less than $20\%$ performance drop. The reason is that LiDAR point clouds are inherently noisy due to the ranging inaccuracy~\cite{carballo2020libre} and self-occlusion, such that the models trained on point clouds are relatively robust to corruptions. 
The results may suggest the indispensability of LiDAR point clouds for reliable 3D object detection. 

\textbf{Comparison of camera-only models.}
Though camera-only detectors are greatly affected by corruptions, we find that multi-view 
methods outperform monocular methods in terms of both clean and corruption accuracy. From Fig.~\ref{fig:plot-nuscene}, the overall performance drop of FCOS3D and PGD is $57\%$ and $54\%$, while that of DETR3D and BEVFormer is $46\%$ and $45\%$, respectively. 
Since monocular methods directly predict 3D objects from single images without considering 3D scene structure, they are more prone to noises \cite{wang2022detr3d} and exhibit inferior performance.
Besides, BEVFormer performs better than DETR3D, especially under object-level corruptions (\eg, \emph{Shear}, \emph{Rotation}), since it can capture both semantic and location information of objects in the BEV space with being less affected by varying object shapes~\cite{li2022delving}. 


\textbf{Comparison of LiDAR-camera fusion models.}
Based on the above analysis, fusion models demonstrate superior corruption robustness on nuScene-C. By carefully examining their performance, we find that there exists a trade-off between robustness under image corruptions and point cloud corruptions. Specifically, FUTR3D suffers from the largest performance drop ($12.9\%$ on average) under \emph{Gaussian}, \emph{Uniform} and \emph{Impluse} noises of images, compared with $2.5\%$ of TransFusion and $5.3\%$ of BEVFusion.
However, under \emph{Motion Compensation} that significantly distorts point clouds, FUTR3D obtains the highest mAP of $31.87\%$ while TransFusion only has $9.01\%$ mAP. 
The reason behind this trade-off is that fusion models have varying reliance on images or point clouds, resulting in the inconsistent robustness under the corresponding corruptions of different sensors.
\section{Discussion and Conclusion}\label{sec:conclusion}

In this paper, we systematically design 27 types of common corruptions in 3D object detection to benchmark corruption robustness of existing 3D object detectors. We establish three corruption robustness benchmarks---KITTI-C, nuScenes-C, and Waymo-C by synthesizing the corruptions on public datasets. By conducting large-scale experiments on 24 diverse 3D object detection models under corruptions, we draw some important findings, as summarized below:
\begin{itemize}[leftmargin=3ex]
\vspace{-1ex}
\item[1)] In general, the corruption robustness of 3D object detection models is largely correlated with their clean performance, similar to the observation in~\cite{hendrycks2018benchmarking}. 
\vspace{-1.2ex}
\item[2)] Among all corruption types, motion-level ones degrade the model performance most, which pose a significant threat to autonomous driving. Weather-level corruptions are also influential to models trained on normal weather.

\vspace{-1.2ex}
\item[3)] Among all 3D detectors, LiDAR-camera fusion models have better corruption robustness, especially under those that apply distortions to only one modality. However, they are also exposed to corruptions from both sensors, leading to degraded performance in this case. Besides, there is a trade-off between robustness under image corruptions and
point cloud corruptions of fusion models.

\vspace{-1.2ex}
\item[4)] Camera-only models are more easily affected by common corruptions, demonstrating the  indispensability of LiDAR point clouds for reliable 3D detection or the necessity of developing more robust camera-only models.

\vspace{-1.2ex}
\item[5)] In Appendix E, we further try several data augmentation strategies, including those applied to point clouds~\cite{choi2021part,zhang2022pointcutmix} and images~\cite{zhang2018mixup,yun2019cutmix}. The experiments validate that they can hardly improve corruption robustness, leaving robustness enhancement of 3D object detection an open problem for future research.
\end{itemize}
\vspace{-0.5ex}

We hope our comprehensive benchmarks, in-depth analyses, and insightful findings can be helpful for understanding the corruption robustness of 3D object detection models and improving their robustness in future.

\section*{Acknowledgement}

This work was supported by the National Key Research and Development Program of China (No. 2017YFA0700904), NSFC Projects (Nos. 62276149, 62061136001, 62076145, 62076147, U19B2034, U1811461, U19A2081, 61972224), Beijing NSF Project (No. JQ19016), BNRist (BNR2022RC01006), Tsinghua Institute for Guo Qiang, and the High Performance Computing Center, Tsinghua University. Y. Dong was also supported by the China National Postdoctoral Program for Innovative Talents and Shuimu Tsinghua Scholar Program. J. Zhu was also supported by the XPlorer Prize.

{\small
\bibliographystyle{ieee_fullname}
\bibliography{arxiv}
}

\clearpage
\appendix

\numberwithin{equation}{section}
\numberwithin{figure}{section}
\numberwithin{table}{section}


\section{More Details of Common Corruptions}

In this section, we provide more technical details about the 27 common corruptions in 3D object detection studied in this paper. 

\subsection{Implementation Details}\label{app:a-1}

First, we introduce the implementation details and hyperparameters of the 27 common corruptions in the three benchmarks---KITTI-C, nuScenes-C, and Waymo-C. Note that we have five severities for each corruption, thus we introduce the corresponding hyperparameter configuration of each severity. 


 
 \textbf{\emph{Snow.}}
For \textbf{LiDAR}, we adopt the method proposed in \cite{hahner2022lidar} to simulate snow on clean data for nuScenes-C, and use LISA \cite{kilic2021lidar} for KITTI-C and Waymo-C. Following the definition in \cite{no2005surface}, we set the \textit{snowfall rate} as \{0.20, 0.73, 1.5625, 3.125, 7.29\} under the five severities for both LISA \cite{kilic2021lidar} and \cite{hahner2022lidar}. For \textbf{Camera}, we use the imgaug library \cite{imgaug} to implement it, and use the pre-defined severities \{1, 2, 3, 4, 5\} to simulate different intensity of snow. To keep the consistency with the STF dataset \cite{bijelic2020seeing}, we add a 30\%-opacity gray mask layer, and reduce the brightness by 30\%.


\textbf{\emph{Rain.}}
For \textbf{LiDAR}, we use the LISA rain simulation method proposed in \cite{hahner2022lidar} for KITTI-C, nuScenes-C, and Waymo-C. Following the real-world rainfall rates defined in\cite{wiki:Rain} and \cite{meteorological2015manobs}, we set the parameter of \textit{rainfall rate} as \{0.20, 0.73, 1.5625, 3.125, 7.29\} in LISA to simulate rain intensity across light rain, moderate rain and heavy rain. For \textbf{Camera}, we set the parameter of \textit{rainfall density} as \{0.01, 0.06, 0.10, 0.15, 0.20\} in RainLayer in imgaug library~\cite{imgaug} to simulate different severity of rain. Besides, we also add a 30\%-opacity gray mask layer, and reduce the brightness by 30\%.

\textbf{\emph{Fog.}} For \textbf{LiDAR}, we use the method proposed in \cite{hahner2021fog} for all three benchmarks. The parameter of $\alpha$ in \cite{hahner2021fog} can represent meteorological optical range in real foggy weather. Following the settings in their paper, we set $\alpha$ to \{0.005, 0.01, 0.02, 0.03, 0.06\} for different severities of fog. For \textbf{Camera}, we use imgaug library \cite{imgaug} to implement it, and use the predefined severities \{1, 2, 3, 4, 5\} to simulate different intensity of fog. Besides, we also add a \{10\%, 20\%, 30\%, 40\%, 50\%\}-opacity gray mask layer.

\textbf{\emph{Strong Sunlight.}}
For \textbf{LiDAR}, following the observations in \cite{carballo2020libre}, we simulate it by adding 2m Gaussian noises to points.  We use the ratio of \{1\%, 2\%, 3\%, 4\%, 5\%\} noisy points to define the severity. For \textbf{Camera}, we set \{30, 40, 50, 60, 70\}-pixel size sun in automold library \cite{UjjwalSaxena2018} for strong sunlight simulation.

 \textbf{\emph{Density Decrease.}}
We randomly delete \{6\%, 12\%, 18\%, 24\%, 30\%\} of points in one frame of LiDAR. 

 \textbf{\emph{Cutout.}}
We randomly remove \{2, 3, 5, 7, 10\} groups of point cloud, where the number of points within each group is $\frac{N}{50}$, and each group is within a ball in the Euclidean space, where $N$ is the total numbers of point in one frame of LiDAR. 

 \textbf{\emph{LiDAR Crosstalk.}}
Following \cite{brinon2021methodology}, we select a subset of points with the ratio of \{0.4\%, 0.8\%, 1.2\%, 1.6\%, 2\%\} to add 3m Gaussian noises.

 \textbf{\emph{FOV Lost.}}
Five groups of FOV lost are selected, the reserved angle range is \{(-105, 105), (-90, 90), (-75, 75), (-60, 60), (-45, 45)\}. 

 \textbf{\emph{Gaussian Noise.}}
For \textbf{LiDAR}, we add Gaussian noises to all points with the severities of \{0.02m, 0.04m, 0.06m, 0.08m, 0.10m\}. For \textbf{Camera}, we use imgaug library \cite{imgaug} to implement it, and use the predefined severities \{1, 2, 3, 4, 5\} to simulate different intensity of \textit{Gaussian Noise}.

 \textbf{\emph{Uniform Noise.}}
For \textbf{LiDAR}, we add uniform noises to all points with the severities of \{0.02m, 0.04m, 0.06m, 0.08m, 0.10m\}. For \textbf{Camera}, we add the uniform noise of $\pm$\{0.08, 0.12, 0.18, 0.26, 0.38\} to the image, thus to simulate different intensity of \textit{Uniform Noise}.

 \textbf{\emph{Impulse Noise.}}
For \textbf{LiDAR}, we select the number of points in $\{\frac{N}{30}, \frac{N}{25}, \frac{N}{20}, \frac{N}{15}, \frac{N}{10}\}$ to add impulse noise and represent the severities, where $N$ is the total numbers of point in one frame of LiDAR. For \textbf{Camera}, we use imgaug library \cite{imgaug} to implement it, and use the predefined severities \{1, 2, 3, 4, 5\} to simulate different intensity of \textit{Impulse Noise}.

 \textbf{\emph{Motion Compensation.}}
We add Gaussian noises to the rotation and translation matrices of the vehicle's ego pose. The noises are \{0.02, 0.04, 0.06, 0.08, 0.10\} for the rotation matrix and \{0.002, 0.004, 0.006, 0.008, 0.010\} for the translation matrix.

 \textbf{\emph{Moving Object.}}
For \textbf{LiDAR}, we first divide a 3D bounding box to three parts, and then move the second part forward with $\frac{c}{2}$, and move the third part forward with $c$, where $c$ is \{0.2, 0.3, 0.4, 0.5, 0.6\}. For \textbf{Camera}, we first use ground-truth 3D bounding boxes to select the object regions and use imgaug library \cite{imgaug} with zoom factor at \{2, 3, 4, 5, 6\} in these regions.

 \textbf{\emph{Motion Blur.}}
We use imgaug library \cite{imgaug} and set the zoom factor to \{2, 3, 4, 5, 6\} to implement it.


 \textbf{\emph{Local Density Decrease.}}
We randomly delete 75\% of points within a group, the group number are \{1, 2, 3, 4, 5\} and each group has 10\% points of a LiDAR frame. 

 \textbf{\emph{Local Cutout.}}
Similar to \textit{cutout}, we randomly remove  \{30\%, 40\%, 50\%, 60\%, 70\%\} of points that within a ball in the Euclidean space, all points are within the objects' 3D bounding boxes.
\begin{table*}[!t]
\begin{center}\footnotesize\setlength{\tabcolsep}{4pt}
    \begin{tabular}{c|ccccc|ccc|cccc}
    \hline
     \multirow{2}{*}{} & \multicolumn{5}{c|}{Corruption Types} & \multicolumn{3}{c|}{Datasets} & \multicolumn{4}{c}{3D Object Detection Models} \\
      & Weather & Sensor & Motion & Object & Alignment & KITTI & nuScenes & Waymo & LiDAR-only & Camera-only & Fusion & \#Models \\ 
     \hline\hline
     Li et al.~\cite{li2022common} & \cmark & \cmark & \xmark & \cmark & \xmark & \cmark & \xmark & \xmark & \cmark & \xmark & \xmark & 7 \\
     Yu et al.~\cite{yu2022benchmarking} & \xmark & \cmark & \xmark & \cmark & \cmark & \xmark & \cmark & \cmark & \cmark & \cmark & \cmark & 10\\
     \textbf{Ours} & \cmark & \cmark & \cmark & \cmark & \cmark & \cmark & \cmark & \cmark & \cmark & \cmark & \cmark & \bf24 \\
    \hline
    \end{tabular}
    \end{center}
    \vspace{-2ex}
    \caption{Comparison between our work and two related works~\cite{li2022common,yu2022benchmarking} in terms of corruption types, datasets, and evaluation models. Our benchmark is more comprehensive in all aspects.}
    \label{tab:compare}
\end{table*}

 \textbf{\emph{Local Gaussian Noise.}}
Similar to the sensor-level Gaussian noise, we add Gaussian noises to the points within the objects' 3D bounding boxes. The noises are \{0.02m, 0.04m, 0.06m, 0.08m, 0.10.\}.

 \textbf{\emph{Local Uniform Noise.}}
 Similar to the sensor-level uniform noise, we add uniform noises to points within the objects' 3D bounding boxes with the severities of \{0.02m, 0.04m, 0.06m, 0.08m, 0.10m\}. 

 \textbf{\emph{Local Impluse Noise.}}
Similar to the sensor-level impluse noise, we select the number of points in \{$\frac{N_{bbox}}{30}$, $\frac{N_{bbox}}{25}$, $\frac{N_{bbox}}{20}$, $\frac{N_{bbox}}{15}$, $\frac{N_{bbox}}{10}$\}  within the objects' 3D bounding boxes to add impulse noises and  represent the severities, where $N_{bbox}$ is the total number of points within the 3D bounding box.

\textbf{\emph{Shear.}}
For \textbf{LiDAR}, we use the shear transformation for the point cloud within the objects' 3D bounding boxes. Let $\mathcal{X}$ represent the point cloud within a 3D bounding box, the transformation can represent as: 
\begin{equation}
\mathcal{X}_t = \mathcal{X}\begin{bmatrix}
1 & 0 & d\\
e & 1 & f\\
g & 0 & 1\\
\end{bmatrix},
\end{equation}
where $d,e,f,g$ are selected from the uniform distribution bounded by $\pm$\{(0.0, 0.1), (0.05, 0.15), (0.1, 0.2), (0.15, (0.20, 0.30)\}.
For \textbf{Camera}, we obtain the 3D bounding box from annotation, do the same shear transformation as in LiDAR points on 8 bounding box corners in the 3D space. Let $ \mathcal{X}_c$ as 8 corners, the transformation can be:
\begin{equation}
\mathcal{X}_{ct} = \mathcal{X}_c\begin{bmatrix}
1 & 0 & d\\
e & 1 & f\\
g & 0 & 1\\
\end{bmatrix}.
\end{equation}
Then we project the corners before and after transformation to images. The projected corners are used as control points to do Thin Plate Spline~(TPS) on images. The severities are the same with LiDAR.

\textbf{\emph{Scale.}}
For \textbf{LiDAR}, we use the scale transformation for the point cloud within the objects' 3D bounding boxes. Let $\mathcal{X}$ represent the point cloud within 3D bounding box, the transformation can represent as: 
\begin{equation}
\mathcal{X}_t = \mathcal{X}\begin{bmatrix}
a & b & c\\
\end{bmatrix},
\end{equation}
where $a,b,c$ are selected from $\pm$\{0.04, 0.08, 0.12, 0.16, 0.20\}. For \textbf{Camera}, we obtain the 3D bounding box from annotation, do the same scaling transformation on 8 bounding box corners in 3D space. Let $ \mathcal{X}_c$ as 8 corners, the transformation can be:
\begin{equation}
\mathcal{X}_{ct} = \mathcal{X}_c\begin{bmatrix}
a & b & c\\
\end{bmatrix}.
\end{equation}
Then we project the corners before and after transformation to images. The projected corners are used as control points to do TPS on images. The severities are the same with LiDAR.

\textbf{\emph{Rotation.}}
For \textbf{LiDAR}, we rotate each 3D bounding box along $z$ axis with angles sampled from the uniform distribution of $\pm$\{(0, 2), (3, 4), (5, 6), (7, 8), (9, 10)\}. For \textbf{Camera}, we obtain the 3D bounding box from annotation, do the same rotation transformation on 8 bounding box corners in 3D space. Then we project the corners before and after transformation to images. The projected corners are used as control points to do TPS on images. The severities are the same with LiDAR.

 \textbf{\emph{Spatial Misalignment.}}
We add Gaussian noises to the calibration matrices between  \textbf{LiDAR} and  \textbf{Camera}. Specifically, the noises are \{0.02, 0.04, 0.06, 0.08, 0.10\} for the rotation matrix and \{0.002, 0.004, 0.006, 0.008, 0.010\} for the translation matrix.

\textbf{\emph{Temporal Misalignment.}}
For \textbf{LiDAR}, the stucked frames are \{2, 4, 6, 8, 10\}. For \textbf{Camera}, the stucked frames also are \{2, 4, 6, 8, 10\}.




\subsection{Comparison with Related Work}

As we mentioned in Sec.~2.2, there are two concurrent works~\cite{li2022common,yu2022benchmarking}, which also study the corruption robustness of 3D object detection in autonomous driving. Compared with them, our benchmark is more comprehensive in terms of corruption types, evaluated datasets, and studied 3D object detection models, as shown in Table~\ref{tab:compare}. Notably, they did not consider motion-level corruptions and we for the first time study motion-level corruptions in a comprehensive robustness benchmark.

\begin{figure*}[t]
  \centering
  \includegraphics[width=0.99\linewidth]{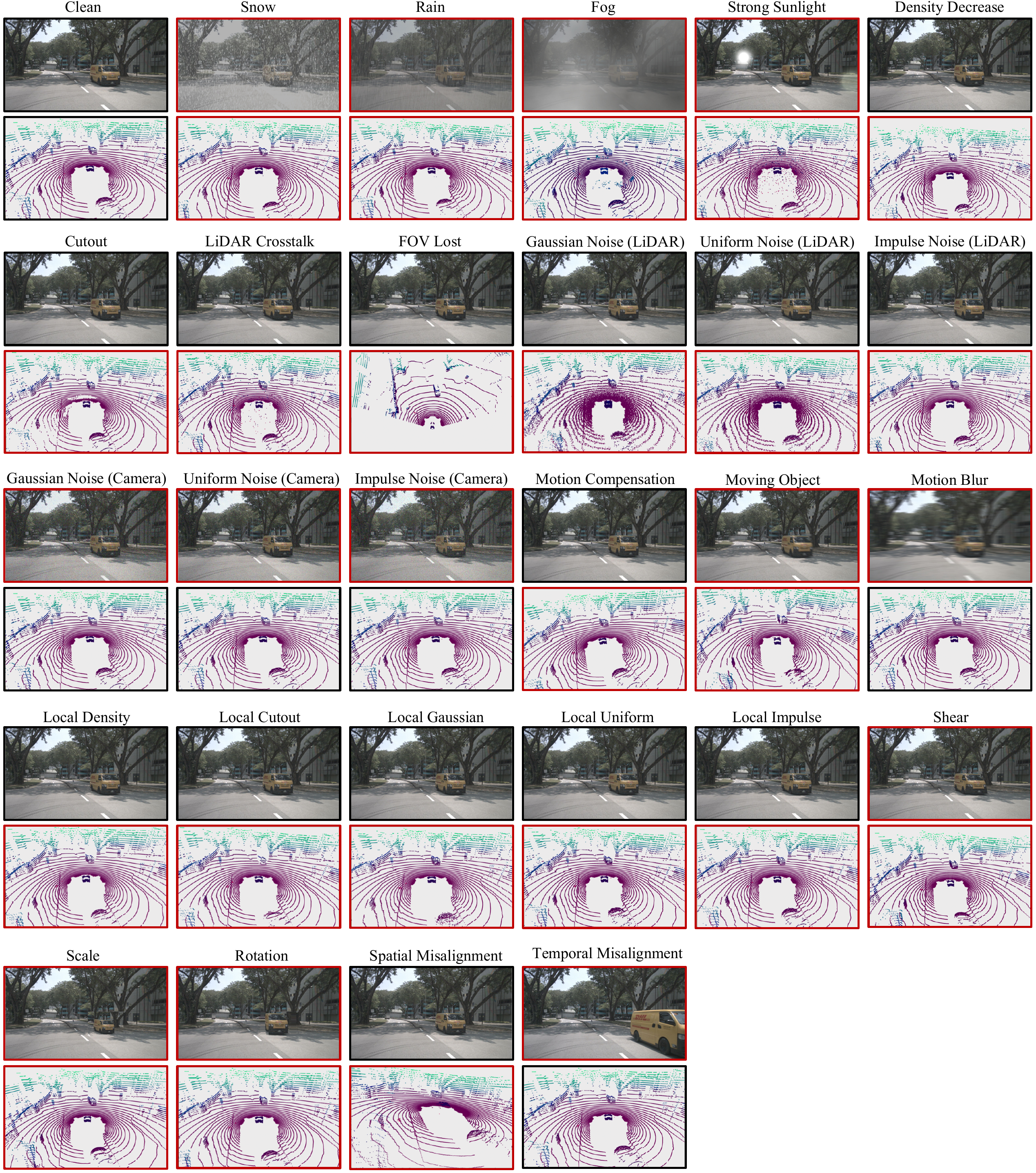}
   \caption{Full visualization results of all corruptions in our benchmark (best viewed when zoomed in). The images or point clouds in {\color{red} red} boxes are modified under the corresponding corruption, while the images or point clouds in black boxes are kept unchanged.}
   \label{fig:vis-full}
\end{figure*}

\subsection{Visualization}

We show the full visualization of all 27 corruptions in Fig.~\ref{fig:vis-full}. Note that an input (image or point cloud) may not be modified under a corruption, thus we mark it by the black box. For input that has been modified under the corruption, we mark it by the {\color{red}red} box. 
 
Since we have 24 corruptions with 5 severities, the KITTI-C dataset is $120\times$ larger than the KITTI validation set, requiring more than 750G storage space. nuScenes-C and Waymo-C are even much bigger than KITTI-C. We will plan to release the full benchmarks.

\begin{table*}[!t]
\footnotesize
\setlength{\tabcolsep}{4.5pt}
\begin{center}
\begin{tabular}{c|c|ccc|cccc|ccc}
\hline
\multicolumn{2}{c|}{\multirow{2}{*}{\textbf{Corruption}}} & \multicolumn{3}{c|}{\textbf{LiDAR-only}} &  \multicolumn{4}{c|}{\textbf{Camera-only}} & \multicolumn{3}{c}{\textbf{LC Fusion}} \\
\multicolumn{2}{c|}{} & PointPillars & SSN & CenterPoint & FCOS3D & PGD & DETR3D & BEVFormer & FUTR3D & TransFusion & BEVFusion \\
\hline\hline
\multirow{4}{*}{\textbf{Synthetic}} & Snow & 27.57 & 46.38 & 55.90 & 2.01 & 2.30 & 5.08 & 5.73 & 52.73 & 63.30 & 62.84 \\
& Rain & 27.71 & 46.50 & 56.08 & 13.00 & 13.51 & 20.39 & 24.97 & 58.40 & 65.35 & 66.13 \\
& Fog & 24.49 & 41.64 & 43.78 & 13.53 & 12.83 & 27.89 & 32.76 & 53.19 & 53.67 & 54.10 \\
& Sunlight & 23.71 & 40.28 & 54.20 & 17.20 & 22.77 & 34.66 & 41.68 & 57.70 & 55.14 & 64.42 \\
\hline
\multirow{4}{*}{\textbf{Real}} & Sunny & 27.60 & 47.01 & 56.00 & 23.88 & 23.09 & 34.33 & 41.02 & 63.67 & 66.24 & 67.20 \\
& Rainy & 27.12 & 44.23 & 54.20 & 23.18 & 22.32 & 36.25 & 43.95 & 65.73 & 66.62 & 68.71 \\
& Day & 27.41 & 46.70 & 56.69 & 24.15 & 23.53 & 34.99 & 41.88 & 64.18 & 66.37 & 67.50 \\
& Night & 18.74 & 24.48 & 30.98 & 12.13 & 11.15 & 16.04 & 21.21 & 38.44 & 41.56 & 39.47 \\
\hline
\end{tabular}
\end{center}
\vspace{-4ex}
\caption{Comparison of model performance under synthetic weathers and real-world dataset of different conditions.}
\label{tab:compare}
\end{table*}

\subsection{Naturalness of Common Corruptions}

\textbf{Quantitative analysis.} In Sec.~3, we have discussed the gap between synthetic and real-world corruptions. Here we further examine the performance of 3D object detection models under adverse weathers crafted by synthetic methods or collected in the real dataset. The nuScenes~\cite{caesar2020nuscenes} dataset has provided the annotations for \emph{Day}, \emph{Night}, \emph{Sunny}, and \emph{Rainy}. Thus we show the performance of 10 3D object detection models introduced in Sec.~4.2 under both synthetic and real-world conditions. Table~\ref{tab:compare} shows the results. Specifically, the model performance under synthetic and real rain weather is largely consistent. The only exception is for camera-only models, where the gap is relatively large. This is due to the difficulty of synthesizing more realistic images under adverse weathers. However, the relative performance of different models is consistent. The results can prove the validity of using our corruption benchmarks for evaluating the robustness of 3D object detection models.

\textbf{Data quality check.} As pointed out by one of the reviewers, data quality check is an important aspect of our benchmark. Actually, we did data quality checks when we simulated the corruptions. We ensured that the objects are detectable for humans by appropriately adjusting the hyperparameters (i.e., severity) of each corruption, as detailed in \cref{app:a-1}. The only exceptions are Cutout and FOV Lost, which may drop objects in the point clouds. We think that a potential solution is to discard the ground-truth objects if they are invisible. However, we found that the evaluations are hard to perform and compare with each other since fusion models have the ability to detect those objects based on accurate camera inputs. Therefore, we tend to keep the original evaluation results (different from what we promise in the rebuttal) and will further consider this problem.

\section{Additional Results on KITTI-C}

\begin{table*}[!t]
\footnotesize
\setlength{\tabcolsep}{2.8pt}
\begin{center}
\begin{tabular}{c|c|cccccc|ccc|cc}
\hline
\multicolumn{2}{c|}{\multirow{2}{*}{\textbf{Corruption}}} & \multicolumn{6}{c|}{\textbf{LiDAR-only}} &  \multicolumn{3}{c|}{\textbf{Camera-only}} & \multicolumn{2}{c}{\textbf{LC Fusion}} \\
\multicolumn{2}{c|}{} & SECOND & PointPillars & PointRCNN & Part-$A^2$ & PV-RCNN & 3DSSD & SMOKE & PGD & ImVoxelNet & EPNet & Focals Conv \\
\hline\hline
\rowcolor{lightgray}\multicolumn{2}{c|}{\textbf{None} ($\mathrm{AP_{clean}}$)}  & 90.53  & 87.75       & 91.65     & 91.68  & \bf92.10    & 91.07 & 10.42 & 12.72 & \bf17.85      & \bf92.29 & 92.00      \\
\hline
\multirow{4}{*}{\textbf{Weather}} &Snow           & 73.05 & 55.99 & 71.93 & 57.56 & \bf73.06 & 42.76 & 3.68  & 0.86  & 0.30  & 48.03 & 53.80 \\
&Rain           & \bf73.31 & 55.17 & 70.79 & 55.77 & 72.37 & 40.39 & 5.66  & 4.85  & 1.77  & 50.93 & 61.44 \\
&Fog            & 85.58 & 74.27 & 85.01 & 79.74&  \bf89.21 & 61.12 & 8.06  & 1.32  & 2.37  & 64.83 & 68.03 \\
&Sunlight       & 88.05 & 67.42 & 64.90 & 84.25 & 87.27 & 21.59 & 8.75  & 10.94 & 15.72 & 81.77 & \bf90.03 \\
\hline
\multirow{9}{*}{\textbf{Sensor}} & Density        & 90.45  & 86.86       & 91.33     & 90.69  & 91.98    & 90.63 & -     & -     & -          & 91.89 & \bf92.14      \\
 & Cutout         & 81.75  & 78.90       & 83.33     & \bf86.13  & 83.40    & 85.06 & -     & -     & -          & 84.17 & 83.84      \\
 & Crosstalk      & 89.63  & 78.51       & 77.38     & 88.58  & 90.52    & 44.35 & -     & -     & -          & 91.30 & \bf92.01      \\
 & Gaussian (L)   & 73.21  & 86.24       & 74.28     & 65.68  & 74.61    & 69.99 & -     & -     & -          & 66.99 & \bf88.56      \\
 & Uniform (L)    & 89.50  & 87.49       & 89.48     & 86.64  & 90.65    & 87.83 & -     & -     & -          & 89.70 & \bf91.77      \\
 & Impulse (L)    & 90.70  & 87.75       & 90.80     & 90.88  & 91.91    & 90.04 & -     & -     & -          & 91.44 & \bf92.10      \\
 & Gaussian (C)   & -      & -           & -         & -      & -        & -     & 2.09  & 2.83  & 3.74       & \bf91.62 & 89.51      \\
 & Uniform (C)    & -      & -           & -         & -      & -        & -     & 3.81  & 5.45  & 7.66       & \bf91.95 & 91.20      \\
 & Impulse (C)    & -      & -           & -         & -      & -        & -     & 2.57  & 1.97  & 3.35       & \bf91.68 & 89.90      \\
\hline
\multirow{2}{*}{\textbf{Motion}} & Moving Obj.    & 62.64  & 58.49       & 59.29     & 64.40  & 63.36    & 62.48 & 2.69  & 4.57  & 9.63       & \bf66.32 & 54.57      \\
 & Motion Blur    & -      & -           & -         & -      & -        & -     & 5.39  & 5.91  & 6.75       & 89.65 & \bf91.56      \\
\hline
\multirow{8}{*}{\textbf{Object}} & Local Density  & 87.74  & 82.90       & 88.37     &   90.30  & 89.60    & \bf90.33 & -     & -     & -          & 89.40 & 89.60      \\
 & Local Cutout   & 81.29  & 75.22       & 83.30     & \bf87.92  & 84.38    & 87.69 & -     & -     & -          & 82.40 & 85.55      \\
 & Local Gaussian & 82.05  & 87.69       & 82.44     & 87.49  & 77.89    & 87.82 & -     & -     & -          & 85.72 & \bf89.78      \\
 & Local Uniform  & 90.11  & 87.83       & 89.30     & 91.22  & 90.63    & 90.57 & -     & -     & -          & 91.32 & \bf91.88      \\
 & Local Impulse  & 90.58  & 87.84       & 90.60     & 91.82  & 91.91    & 90.89 & -     & -     & -          & 91.67 & \bf92.02      \\
 & Shear          & 47.80  & 45.06       & 45.52     & 37.86  & \bf52.39    & 32.54 & 2.41  & 4.46  & 1.72       & 45.23 & 48.90      \\
 & Scale          & 81.84  & 80.57       & 81.41     & 86.80  & 85.14    & \bf87.31 & 0.12  & 0.14  & 0.39       & 80.53 & 78.82      \\
 & Rotation       & 87.39  & 83.61       & 87.09     & 88.38  & \bf89.29    & 88.71 & 1.43  & 3.19  & 3.68       & 86.70 & 87.02      \\
\hline
\multirow{1}{*}{\textbf{Alignment}} & Spatial        & -      & -           & -         & -      & -        & -     & -     & -     & -          & 42.23 & \bf51.21 \\
\hline
\rowcolor{lightgray}\multicolumn{2}{c|}{\textbf{Average} ($\mathrm{AP_{cor}}$)} & 81.40 & 76.20 & 79.29 & 79.58 & \bf82.61 & 71.16 & 3.89 & 3.87 & \bf4.76 & 78.64 & \bf81.05 \\
\hline
\end{tabular}
\end{center}
\vspace{-4ex}
\caption{The benchmarking results of 11 3D object detectors on \textbf{KITTI-C} based on the car class at easy difficulty.}
\label{tab:results-kitti-car-easy}
\end{table*}

\begin{table*}[!t]
\footnotesize
\setlength{\tabcolsep}{2.8pt}
\begin{center}
\begin{tabular}{c|c|cccccc|ccc|cc}
\hline
\multicolumn{2}{c|}{\multirow{2}{*}{\textbf{Corruption}}} & \multicolumn{6}{c|}{\textbf{LiDAR-only}} &  \multicolumn{3}{c|}{\textbf{Camera-only}} & \multicolumn{2}{c}{\textbf{LC Fusion}} \\
\multicolumn{2}{c|}{} & SECOND & PointPillars & PointRCNN & Part-$A^2$ & PV-RCNN & 3DSSD & SMOKE & PGD & ImVoxelNet & EPNet & Focals Conv \\
\hline\hline
\rowcolor{lightgray}\multicolumn{2}{c|}{\textbf{None} ($\mathrm{AP_{clean}}$)}  & 78.57  & 75.19       & 78.06     & 80.22  & \bf82.49    & 78.23 & 5.57  & 6.20 & \bf9.20       & 80.16 & \bf83.36      \\
\hline
\multirow{4}{*}{\textbf{Weather}} & Snow           & \bf48.62  & 32.96       & 45.41     & 40.03  & 48.62    & 23.15 & 1.92  & 0.44 & 0.20       & 32.39 & 30.41      \\
 & Rain           & \bf48.79  & 32.65       & 45.78     & 39.09  & 48.20    & 22.56 & 3.16  & 2.26 & 0.99       & 34.69 & 35.71      \\
 & Fog            & 68.93  & 58.19       & 68.05     & 68.39  & \bf75.05    & 41.21 & 4.56  & 0.63 & 1.03       & 38.12 & 39.50      \\
 & Sunlight       & 74.62  & 58.69       & 61.11     & 73.55  & 78.02    & 24.70 & 4.91  & 5.42 & 8.24       & 66.43 & \bf78.06      \\
\hline
\multirow{9}{*}{\textbf{Sensor}} & Density        & 77.04  & 72.85       & 77.58     & 78.33  & 81.15    & 74.56 & -     & -    & -          & 79.77 & \bf82.38      \\
 & Cutout         & 70.79  & 67.32       & 71.57     & 73.91  & 74.60    & 70.52 & -     & -    & -          & 73.95 & \bf76.69      \\
 & Crosstalk      & 76.92  & 67.51       & 69.41     & 77.26  & 80.98    & 43.67 & -     & -    & -          & 79.54 & \bf83.22      \\
 & Gaussian (L)   & 61.09  & 71.12       & 56.73     & 58.71  & 62.70    & 55.61 & -     & -    & -          & 56.88 & \bf77.15      \\
 & Uniform (L)    & 75.61  & 74.09       & 72.25     & 75.02  & 78.93    & 71.77 & -     & -    & -          & 75.92 & \bf81.62      \\
 & Impulse (L)    & 78.33  & 74.65       & 76.88     & 78.78  & 81.79    & 75.37 & -     & -    & -          & 79.14 & \bf83.28      \\
 & Gaussian (C)   & -      & -           & -         & -      & -        & -     & 1.18  & 1.26 & 1.96       & 78.20 & \bf79.01      \\
 & Uniform (C)    & -      & -           & -         & -      & -        & -     & 2.19  & 2.46 & 3.90       & 79.14 & \bf81.39      \\
 & Impulse (C)    & -      & -           & -         & -      & -        & -     & 1.52  & 0.82 & 1.71       & 78.51 & \bf78.87      \\
\hline
\multirow{2}{*}{\textbf{Motion}} & Moving Obj.    & 48.02  & 45.47       & 46.23     & \bf53.06  & 50.75    & 50.86 & 1.40  & 1.97 & 4.63       & 50.97 & 45.34      \\
 & Motion Blur    &        &             &           &        &          &       & 2.95  & 2.44 & 3.32       & 72.49 & \bf77.75      \\
\hline
\multirow{8}{*}{\textbf{Object}} & Local Density  & 71.45  & 65.70       & 71.09     & \bf77.58  & 75.39    & 75.05 & -     & -    & -          & 74.36 & 77.30      \\
 & Local Cutout   & 63.25  & 56.69       & 63.50     & \bf72.86  & 68.58    & 70.73 & -     & -    & -          & 66.53 & 72.40      \\
 & Local Gaussian & 68.16  & 73.11       & 65.65     & 75.32  & 68.03    & 72.84 & -     & -    & -          & 72.71 & \bf78.52      \\
 & Local Uniform  & 76.67  & 74.68       & 74.37     & 78.47  & 80.17    & 76.31 & -     & -    & -          & 78.85 & \bf81.99      \\
 & Local Impulse  & 78.47  & 75.18       & 77.38     & 79.98  & 82.33    & 76.91 & -     & -    & -          & 79.79 & \bf83.20      \\
 & Shear          & 39.99  & 38.11       & 38.12     & 37.12  & \bf47.06    & 24.87 & 1.39  & 2.31 & 1.18       & 40.62 & 44.25      \\
 & Scale          & 70.03  & 67.22       & 68.55     & 73.74  & \bf74.89    & 72.56 & 0.12  & 0.15 & 0.32       & 65.68 & 66.65      \\
 & Rotation       & 73.24  & 69.24       & 72.32     & 75.33  & \bf78.02    & 74.35 & 0.84  & 1.67 & 2.18       & 71.91 & 75.25      \\
\hline
\multirow{1}{*}{\textbf{Alignment}} & Spatial        & -      & -           & -         & -      & -        & -     & -     & -    & -          & 33.94 & \bf41.06  \\
\hline
\rowcolor{lightgray}\multicolumn{2}{c|}{\textbf{Average} ($\mathrm{AP_{cor}}$)} & 66.84 & 61.86 & 64.31 & 67.71 & \bf70.28 & 57.77 & 2.18 & 1.82 & \bf2.47 & 65.02 & \bf68.79 \\
\hline
\end{tabular}
\end{center}
\vspace{-4ex}
\caption{The benchmarking results of 11 3D object detectors on \textbf{KITTI-C} based on the car class at hard difficulty.}
\label{tab:results-kitti-car-hard}
\end{table*}

\begin{table*}[!t]
\footnotesize
\begin{center}
\begin{tabular}{c|c|cccc|cc}
\hline
\multicolumn{2}{c|}{\multirow{2}{*}{\textbf{Corruption}}} & \multicolumn{4}{c|}{\textbf{LiDAR-only}} &  \multicolumn{2}{c}{\textbf{Camera-only}} \\
\multicolumn{2}{c|}{} & SECOND & PointPillars & PointRCNN & PV-RCNN & SMOKE & PGD  \\
\hline\hline
\rowcolor{lightgray}\multicolumn{2}{c|}{\textbf{None} ($\mathrm{AP_{clean}}$)}  & 51.14  & 51.41       & 54.40     & \bf54.49    & \bf3.19  & 1.27 \\
\hline
\multirow{4}{*}{\textbf{Weather}} & Snow           & 49.68  & 49.07       & \bf55.73     & 53.01    & 1.11  & 0.14\\
 & Rain           & 50.34  & 49.23       & \bf56.08     & 54.98    & 2.68  & 0.74 \\
 & Fog            & \bf3.10   & 0.05        & 0.14      & 0.67     & 2.71  & 0.22  \\
 & Sunlight       & \bf49.63  & 29.34       & 33.49     & 42.19    & 2.35  & 1.15 \\
\hline
\multirow{9}{*}{\textbf{Sensor}} & Density        & 50.67  & 50.08       & 54.84     & \bf55.59    & -     & -   \\
 & Cutout         & 44.92  & 44.94       & \bf49.38     & 48.05    & -     & -   \\
 & Crosstalk      & \bf50.28  & 38.15       & 43.02     & 48.20    & -     & -   \\
 & Gaussian (L)   & 24.82  & \bf40.00       & 25.89     & 26.32    & -     & -   \\
 & Uniform (L)    & 41.37  & \bf49.54       & 44.24     & 45.58    & -     & -   \\
 & Impulse (L)    & 50.33  & 51.22       & 50.19     & \bf52.39    & -     & -  \\
 & Gaussian (C)   & -      & -           & -         & -        & \bf0.79  & 0.22  \\
 & Uniform (C)    & -      & -           & -         & -        & \bf1.62  & 0.58  \\
 & Impulse (C)    & -      & -           & -         & -        & \bf0.99  & 0.10  \\
\hline
\multirow{2}{*}{\textbf{Motion}} & Moving Obj.    & 3.57   & 3.30        & \bf4.86      & 4.80     & 0.69  & 0.59  \\
 & Motion Blur    & -      & -           & -         & -        & \bf1.19  & 0.82\\
\hline
\multirow{8}{*}{\textbf{Object}} & Local Density  & 37.30  & 33.94       & \bf45.11     & 37.74    & -     & -   \\
 & Local Cutout   & 21.35  & 23.71       & 19.99     & \bf23.96    & -     & -   \\
 & Local Gaussian & 27.49  & \bf43.60       & 28.54     & 29.11    & -     & -  \\
 & Local Uniform  & 44.63  & \bf51.94       & 46.17     & 47.83    & -     & -   \\
 & Local Impulse  & 50.76  & 52.20       & 52.40     & \bf53.20    & -     & -   \\
 & Shear          & 35.91  & 38.31       & 38.52     & \bf38.70    & 0.93  & 0.41  \\
 & Scale          & 46.00  & 46.11       & \bf51.30     & 50.26    & 0.18  & 0.09  \\
 & Rotation       & 50.83  & 51.05       & 54.10     & \bf54.49    & 1.78  & 0.74 \\
\hline
\rowcolor{lightgray}\multicolumn{2}{c|}{\textbf{Average} ($\mathrm{AP_{cor}}$)} & 38.58 & 39.25 & 39.68 & \bf40.37 & 1.42 & 0.48 \\
\hline
\end{tabular}
\end{center}
\vspace{-4ex}
\caption{The benchmarking results of 6 3D object detectors on \textbf{KITTI-C} based on the pedestrian class at moderate difficulty.}
\label{tab:results-kitti-pedestrian}
\end{table*}

\begin{table*}[!t]
\footnotesize
\begin{center}
\begin{tabular}{c|c|cccc|cc}
\hline
\multicolumn{2}{c|}{\multirow{2}{*}{\textbf{Corruption}}} & \multicolumn{4}{c|}{\textbf{LiDAR-only}} &  \multicolumn{2}{c}{\textbf{Camera-only}} \\
\multicolumn{2}{c|}{} & SECOND & PointPillars & PointRCNN & PV-RCNN & SMOKE & PGD  \\
\hline\hline
\rowcolor{lightgray}\multicolumn{2}{c|}{\textbf{None} ($\mathrm{AP_{clean}}$)}   & 66.74  & 62.81       & \bf71.00     & 70.38    & 0.25  & \bf0.86 \\
\hline
\multirow{4}{*}{\textbf{Weather}} & Snow           & 51.35  & 44.15       & \bf57.88     & 55.56    & 0.19  & 0.02 \\
 & Rain           & 51.49  & 44.65       & \bf58.64     & 56.19    & 0.15  & 0.21  \\
 & Fog            & \bf10.91  & 2.77        & 4.29      & 4.31     & 0.30  & 0.03 \\
 & Sunlight       & 61.12  & 45.05       & 60.33     & \bf61.58    & 0.40  & 0.41 \\
\hline
\multirow{9}{*}{\textbf{Sensor}} & Density        & 63.00  & 60.60       & \bf69.66     & 67.76    & -     & -   \\
 & Cutout         & 59.03  & 55.80       & \bf63.46     & 62.28    & -     & -  \\
 & Crosstalk      & 64.02  & 53.52       & 65.25     & \bf67.67    & -     & -  \\
 & Gaussian (L)   & 48.03  & 52.62       & \bf54.08     & 47.53    & -     & -  \\
 & Uniform (L)    & 62.56  & 60.58       & \bf66.77     & 66.40    & -     & -  \\
 & Impulse (L)    & 64.34  & 62.28       & \bf70.13     & 68.69    & -     & -  \\
 & Gaussian (C)   & -      & -           & -         & -        & 0.04  & \bf0.09 \\
 & Uniform (C)    & -      & -           & -         & -        & 0.17  & \bf0.21  \\
 & Impulse (C)    & -      & -           & -         & -        & \bf0.07  & 0.02 \\
\hline
\multirow{2}{*}{\textbf{Motion}} & Moving Obj.    & 21.54  & 21.04       & 23.88     & \bf28.77    & 0.07  & 0.04 \\
 & Motion Blur    & -      & -           & -         & -        & \bf0.17  & 0.08\\
\hline
\multirow{8}{*}{\textbf{Object}} & Local Density  & 47.26  & 36.98       & \bf52.49     & 49.76    & -     & -   \\
 & Local Cutout   & 24.59  & 20.47       & 25.93     & \bf27.01    & -     & -  \\
 & Local Gaussian & 53.61  & 58.94       & \bf60.81     & 56.39    & -     & - \\
 & Local Uniform  & 63.18  & 61.58       & \bf68.80     & 68.30    & -     & -  \\
 & Local Impulse  & 65.11  & 62.79       & \bf70.80     & 69.91    & -     & -  \\
 & Shear          & 57.09  & 56.40       & \bf64.42     & 60.83    & 0.10  & 0.14 \\
 & Scale          & 64.02  & 60.46       & 67.31     & \bf68.30    & 0.08  & 0.03 \\
 & Rotation       & 64.23  & 62.75       & \bf69.67     & 69.39    & 0.08  & 0.16 \\
\hline
\rowcolor{lightgray}\multicolumn{2}{c|}{\textbf{Average} ($\mathrm{AP_{cor}}$)} & 52.45 & 48.60 & \bf56.56 & 55.61 & 0.15 & 0.12  \\
\hline
\end{tabular}
\end{center}
\vspace{-4ex}
\caption{The benchmarking results of 6 3D object detectors on \textbf{KITTI-C} based on the cyclist class at moderate difficulty.}
\label{tab:results-kitti-cyclist}
\end{table*}

\begin{figure*}[t]
  \centering
  \includegraphics[width=0.9\linewidth]{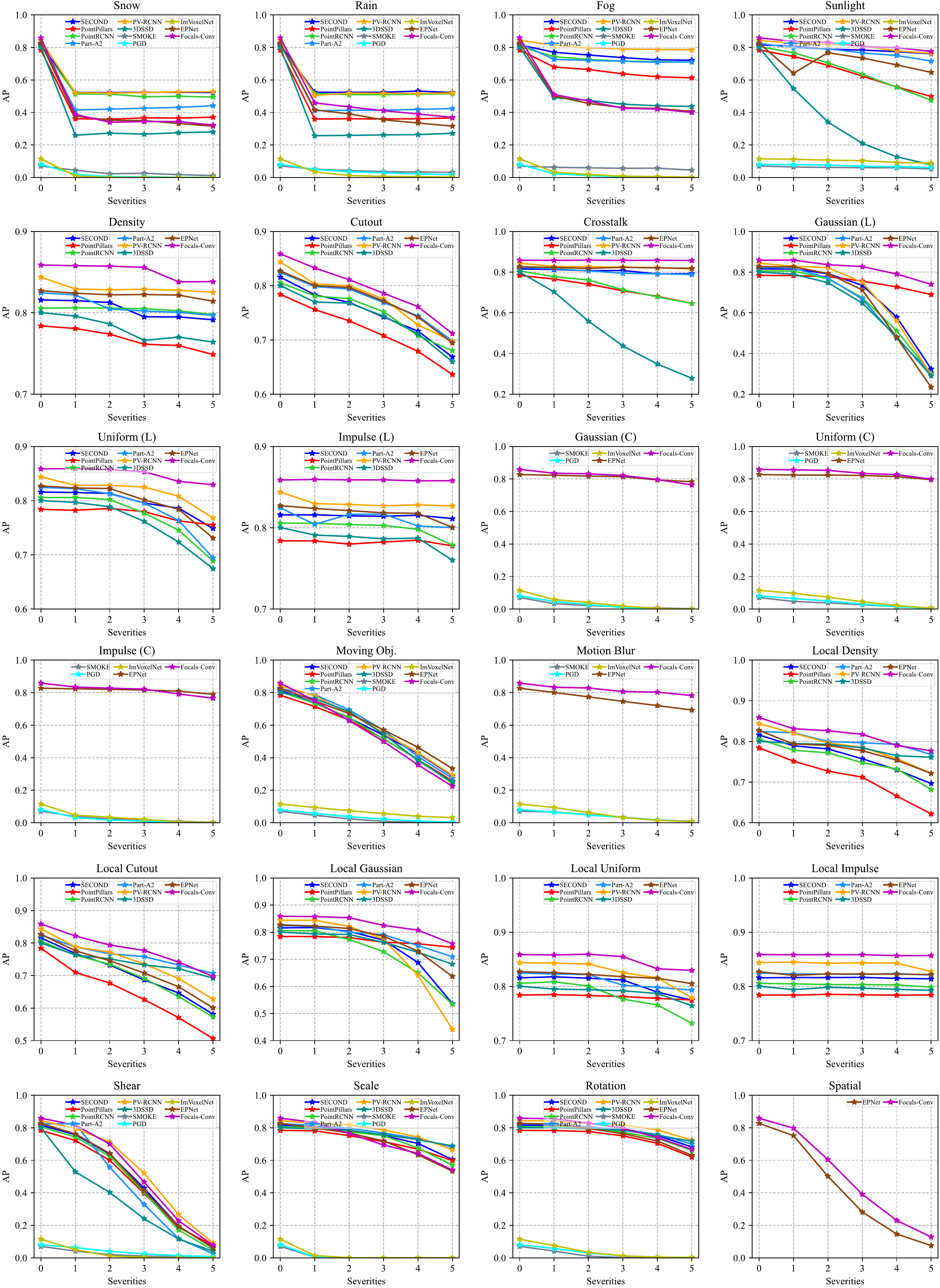}
   \caption{Model performance w.r.t. severity of each corruption on \textbf{KITTI-C}. The results are evaluated based on the car class at moderate difficulty.}
   \label{fig:plot-kitti-severity}
\end{figure*}

In addition to the experimental results in Sec.~5.1, we further provide more results on KITTI-C for other classes and difficulties. We show the corruption robustness of 11 3D object detectors on the car class at easy and hard difficulties in Table~\ref{tab:results-kitti-car-easy} and Table~\ref{tab:results-kitti-car-hard}, respectively. The results are highly consistent with those based on the car class at moderate difficulty in Table 3. For the other two classes (\ie, pedestrian, cyclist), there are only 6 models that can predict these two classes, including SECOND, PointPillars, PointRCNN, PV-RCNN, SMOKE, and PGD. We show the corruption robustness of these 6 3D object detectors on the pedestrian and cyclist classes at the moderate diffucilty in Table~\ref{tab:results-kitti-pedestrian} and Table~\ref{tab:results-kitti-cyclist}, respectively. 
Fig.~\ref{fig:plot-kitti-severity} further shows the model performance under different severities of each corruption. It can be seen that for most corruptions, model performance drops along with the increasing severity.

\section{Additional Results on nuScenes-C}

\begin{table*}[!t]
\footnotesize
\setlength{\tabcolsep}{3.5pt}
\begin{center}
\begin{tabular}{c|c|ccc|cccc|ccc}
\hline
\multicolumn{2}{c|}{\multirow{2}{*}{\textbf{Corruption}}} & \multicolumn{3}{c|}{\textbf{LiDAR-only}} &  \multicolumn{4}{c|}{\textbf{Camera-only}} & \multicolumn{3}{c}{\textbf{LC Fusion}} \\
\multicolumn{2}{c|}{} & PointPillars & SSN & CenterPoint & FCOS3D & PGD & DETR3D & BEVFormer & FUTR3D & TransFusion & BEVFusion \\
\hline\hline
\rowcolor{lightgray}\multicolumn{2}{c|}{\textbf{None} ($\mathrm{NDS_{clean}}$)} & 46.86 & 58.24 & \bf67.33 & 34.69 & 35.04 & 42.23 & \bf51.74 & 68.05 & 69.82 & \bf71.40 \\
\hline
\multirow{4}{*}{\textbf{Weather}} & Snow & 46.67 & 58.07 & 64.92 & 8.57 & 9.83 & 15.53 & 15.61 & 61.52 & 68.29 & \bf68.33\\
& Rain & 46.79 & 58.16 & 64.98 & 26.31 & 26.96 & 31.60& 38.82 & 64.67 & 69.40 & \bf70.14\\
& Fog & 44.91 & 55.42 & 58.11 & 26.05 & 25.83 & 37.26& 45.42 & 61.20 & 62.62 & \bf62.73\\
& Sunlight & 44.57 & 54.59 & 64.41 & 29.34 & 34.77 & 42.20& 51.70& 63.61 & 61.36 & \bf68.95 \\
\hline
\multirow{10}{*}{\textbf{Sensor}} & Density & 46.62 & 57.93 & 66.84 & - & - & - & - & 67.58 & 69.42 & \bf71.01 \\
& Cutout & 44.74 & 55.06 & 65.73 & - & - & - & - & 66.91 & 68.30 & \bf70.09 \\
& Crosstalk & 45.93 & 56.72 & 65.83 & - & - & - & - & 67.17 & 68.83 & \bf70.72 \\
& FOV Lost & 35.69 & 41.61 & 47.07 & - & - & - & - & 45.66 & 47.89 & \bf48.65\\
& Gaussian (L) & 40.62 & 53.24 & 58.08 & - & - & - & - & 64.10 & 62.32 & \bf65.99 \\
& Uniform (L) & 45.44 & 57.03 & 65.22 & - & - & - & - & 67.28 & 68.68 & \bf70.18 \\
& Impulse (L) & 46.21 & 57.42 & 66.22 & - & - & - & - & 67.47 & 69.06 & \bf70.63 \\
& Gaussian (C) & - & - & - & 11.16 & 12.73 &  26.38& 29.60 & 62.92 & 68.94 & \bf69.35\\
& Uniform (C) & - & - & - & 19.55 & 20.63 & 32.13& 37.57 & 64.43 & 69.33 & \bf70.06\\
& Impulse (C) & - & - & - & 11.71 & 12.07 & 26.03& 29.24 & 63.07 & 68.89 & \bf69.25\\
\hline
\multirow{3}{*}{\textbf{Motion}} & Compensation & 20.64 & 27.93 & 27.71 & - & - & - & - & \bf39.62 & 25.69 & 36.76\\
& Moving Obj. & 39.23 & 49.19 & 55.45 & 23.57 & 24.33 &  28.17& 34.59 & 56.41 & \bf60.03 & 59.42\\
& Motion Blur & - & - & - & 23.04 & 23.50 & 23.49& 29.17 & 63.44 & 68.85 & \bf69.38\\
\hline
\multirow{8}{*}{\textbf{Object}} & Local Density & 46.27 & 57.63 & 66.22 & - & - & - & - & 67.62 & 69.34 & \bf70.77\\
& Local Cutout & 39.37 & 48.64 & 60.40 & - & - & - & - & 66.45 & 67.97 & \bf68.11\\
& Local Gaussian & 45.31 & 56.41 & 61.27 & - & - & - & - & 66.85 & 67.96 & \bf68.32\\
& Local Uniform & 46.87 & 58.42 & 66.22 & - & - & - & - & 67.92 & 69.67 & \bf70.68\\
& Local Impulse & 46.93 & 58.41 & 66.70 & - & - & - & - & 67.89 & 69.64 & \bf70.93\\
& Shear & 45.34 & 55.44 & 54.02 & 29.34 & 40.65 & 28.74& 38.77 & 61.15 & \bf66.43 & 62.95 \\
& Scale & 46.58 & 57.85 & 61.27 & 21.68 & 21.41 & 25.48 & 32.81 & 62.00 & \bf67.81 & 66.00 \\
& Rotation & 46.78 & 58.18 & 58.19 & 29.38 & 29.82 &  36.39& 45.45 & 63.67 & \bf67.42 & 66.31 \\
\hline
\multirow{2}{*}{\textbf{Alignment}} & Spatial & - & - & - & - & - & - & - & 67.75 & 69.72 & \bf71.35 \\
& Temporal & - & - & - & - & - & - & - & \bf57.91 & 54.23 & 56.52\\
\hline
\rowcolor{lightgray}\multicolumn{2}{c|}{\textbf{Average} ($\mathrm{NDS_{cor}}$)} &43.41 &53.97 &\bf60.23 &21.64 &22.63 & 29.45 &\bf35.73 &62.82 &64.74 & \bf66.06\\
\hline
\end{tabular}
\end{center}
\vspace{-4ex}
\caption{The benchmarking results of 10 3D object detectors on \textbf{nuScenes-C} under the NDS metric.}
\label{tab:results-nuscenes-nds}
\end{table*}

\begin{figure*}[t]
  \centering
  \includegraphics[width=0.98\linewidth]{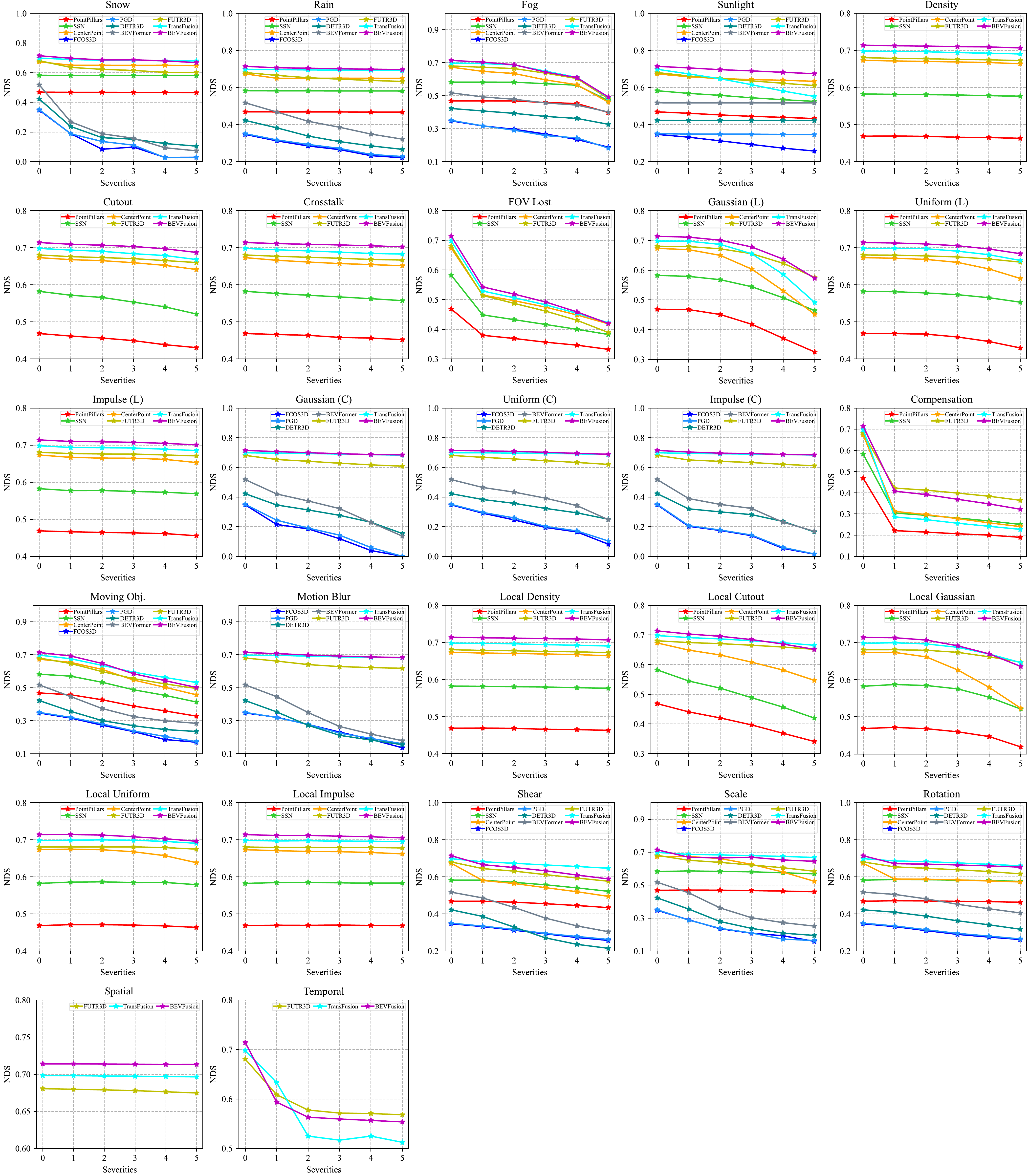}
   \caption{Model performance w.r.t. severity of each corruption on \textbf{nuScenes-C} under the NDS metric.}
   \label{fig:plot-nuscenes-severity}
\end{figure*}

We further provide the results on nuScenes-C under the NDS metric in Table~\ref{tab:results-nuscenes-nds}. The findings are consistent across both the mAP and NDS metrics. We similarly provide the curves of model performance along with severity of each corruption in Fig.~\ref{fig:plot-nuscenes-severity}.

\section{Results on Waymo-C}

\begin{table}[h]
\footnotesize
\setlength{\tabcolsep}{3pt}
\begin{center}
\begin{tabular}{c|c|c|c|c}
\hline
\multicolumn{2}{c|}{\textbf{Corruption}}  & PointPillars  &BEVFormer & TransFusion \\
\hline\hline
\rowcolor{lightgray}\multicolumn{2}{c|}{\textbf{None} ($\mathrm{L2/mAPH_{clean}}$)} & 59.13 & 22.96 & \bf60.16  \\
\hline
\multirow{4}{*}{\textbf{Weather}} & Snow & 33.39 & 5.98 & \bf34.98  \\
& Rain & 34.85 & 12.61 & \bf37.20  \\
& Fog & 0.92 & \bf18.50 & 0.81  \\
& Sunlight & 30.00 & 23.50 & \bf47.10  \\
\hline
\multirow{9}{*}{\textbf{Sensor}} & Density & 58.33 & - & \bf58.95  \\
& Cutout  & 54.14 & - & \bf55.40 \\
& Crosstalk  & 43.04 & - & \bf57.06 \\
& FOV Lost & 26.16 & - & \bf26.28 \\
& Gaussian (L) & \bf52.27 & - & 36.68 \\
& Uniform (L) & \bf57.85 & - & 56.54 \\
& Impulse (L) & 58.76 & - & \bf59.49 \\
& Gaussian (C) & - & 14.64 & \bf59.98 \\
& Uniform (C) & - & 16.00 & \bf60.01 \\
& Impulse (C) & - & 14.12 & \bf59.95 \\
\hline
\multirow{2}{*}{\textbf{Motion}} & Moving Obj. & 40.02 & 12.99 & \bf41.55 \\
& Motion Blur & - & 7.98 & \bf59.29 \\
\hline
\multirow{8}{*}{\textbf{Object}} & Local Density & 58.45 & - & \bf59.39 \\
& Local Cutout & 56.74 & - & \bf58.11 \\
& Local Gaussian & 57.88 & - & \bf58.23 \\
& Local Uniform & 58.77 & - & \bf59.68 \\
& Local Impulse & 58.85 & - & \bf59.95 \\
& Shear & 49.61 & 6.93 & \bf52.73 \\
& Scale & 54.25 & 1.89 & \bf55.62 \\
& Rotation & 54.89 & 10.36 & \bf58.04  \\
\hline
\multirow{1}{*}{\textbf{Alignment}} & Spatial & - & - & \bf59.66 \\
\hline
\rowcolor{lightgray}\multicolumn{2}{c|}{\textbf{Average} ($\mathrm{L2/mAPH_{cor}}$)} & 46.96 & 12.13 & \bf50.91 \\
\hline
\end{tabular}
\end{center}
\vspace{-4ex}
\caption{The benchmarking results of 3 3D object detectors on \textbf{Waymo-C}. We show the performance under each corruption and the overall corruption robustness mAPcor averaged over all corruption types.}
\label{tab:results-waymo}
\end{table}

We evaluate the corruption robustness of PointPillars \cite{lang2019pointpillars}, BEVFormer \cite{li2022bevformer}, and TransFusion \cite{bai2022transfusion} on Waymo-C in Table~\ref{tab:results-waymo}.
Since we do not have enough models for more comprehensive comparison, we can only draw the conclusion that the LiDAR-camera fusion model TransFusion demonstrates better performance than the other models. We would continuously evaluate more 3D object detection models on Waymo-C in future. 

\section{Data Augmentation as Potential Defense}

\begin{table}[!t]
\begin{center}\small
    \begin{tabular}{ccc}
    \hline
       Augmentation & Second & PV-RCNN  \\
     \hline
     None & 70.45 & 72.59  \\
     \hline
      PA-AUG & 70.63 & 65.93  \\
      Dropout & 71.29 & 67.91  \\
     PointCutMix-R & 68.97 & 64.43  \\
    
    \hline
    \end{tabular}
    \end{center}
    \vspace{-4ex}
    \captionsetup{font={small}}
    \caption{The corruption robustness ($\mathrm{AP_{cor}}$) of SECOND and PV-RCNN with different data augmentations on \textbf{KITTI-C}.}
    \label{pl_aug}
\end{table}

\begin{table}[!t]
\begin{center}\small
    \begin{tabular}{cc|c}
    \hline
       Image Aug. & Point Cloud Aug. & Focals Conv  \\
     \hline
     None & None & 71.87 \\
     \hline
     Mixup & PA-AUG & 55.32 \\
     Mixup & Dropout & 25.07 \\
     Mixup & PointCutMix-R & 48.12 \\
     \hline
     CutMix & PA-AUG & 30.01 \\
     CutMix & Dropout & 53.90 \\
     CutMix &  PointCutMix-R & 28.83  \\
     \hline
    \end{tabular}
    \end{center}
    \vspace{-4ex}
    \captionsetup{font={small}}
    \caption{The corruption robustness ($\mathrm{AP_{cor}}$) of Focals Conv with different data augmentations on \textbf{KITTI-C}.}
    \label{pcfusion_aug}
\end{table}


In this section, we explore data augmentation for improving the robustness of 3D object detection models under common corruptions. We adopt the PA-AUG and Dropout \cite{choi2021part} methods and PointCutMix-R~\cite{zhang2022pointcutmix} method for LiDAR point cloud augmentation. For the camera modality, we use two famous image data augmentations, which are Mixup \cite{zhang2018mixup} and CutMix \cite{yun2019cutmix}.

\textbf{For LiDAR-only models.} We perform experiments on SECOND~\cite{yan2018second} and PV-RCNN~\cite{shi2020pv} due to their superior robustness among all LiDAR-only models. The results are shown in Table \ref{pl_aug}.  These augmentations do not improve performance consistently. The Dropout augmentation only improves the corruption robustness of SECOND by 0.84. But these augmentations drop the robustness of PV-RCNN by more than 4.68. The reason is that these augmentations degrade model performance on clean data. Since the corruption robustness is highly correlated with clean performance, the effectiveness of these augmentations is limited.

\textbf{For LiDAR-camera fusion models.} We choose Focals Conv~\cite{chen2022focal} as the target to study the effectiveness of data augmentation techniques.  The multi-modal data augmentation is still an open question in computer vision community\cite{zhang2022cat}, especially in 3D object detection. Here, we explore the synergistic data augmentation of camera modalities and LiDAR modalities. Specifically, we choose three point cloud augmentations and two image augmentations for LiDAR-camera fusion models. The results are shown in Table \ref{pcfusion_aug}. It can be seen that the combination of data augmentations of both modalities degrades the performance a lot. Therefore, it remains an open problem of improving the corruption robustness of 3D object detectors, especially LiDAR-camera fusion models.

\end{document}